\documentclass{article}

\usepackage{arxiv}

\usepackage[utf8]{inputenc} 
\usepackage[T1]{fontenc}    
\usepackage{hyperref}       
\usepackage{url}            
\usepackage{booktabs}       
\usepackage{amsfonts}       
\usepackage{nicefrac}       
\usepackage{microtype}      
\usepackage{graphicx}
\usepackage{amssymb}
\usepackage{amsmath}
\usepackage{amsthm}
\usepackage{array}
\usepackage{booktabs}
\usepackage{multirow}
\usepackage{siunitx}
\usepackage{pifont}
\usepackage{textcomp}
\usepackage{stfloats}
\usepackage{balance}
\usepackage{longtable}
\usepackage{pdflscape}
\usepackage[numbers,square]{natbib}
\usepackage[caption=false,font=normalsize,labelfont=sf,textfont=sf]{subfig}

\newcommand{\cmark}{\ding{51}}
\newcommand{\xmark}{\ding{55}}
\newcommand{\figpanels}[2]{Fig.~\ref{#1}(#2)}
\newcommand{\Eqref}[1]{(Eq.~\ref{#1})}
\newcommand{\Eqsref}[2]{(Eq.~\ref{#1} and~\ref{#2})}
\hypersetup{hidelinks}

\title{SEAGAN: domain-Specific and Edge-Aware Graph Attention Network for Dynamic Plant Processes}

\author{
 Antriksh Srivastava \\
  Center of Studies in Resources Engineering\\
  Indian Institute of Technology Bombay\\
  Mumbai, India 400076 \\
  \texttt{srivastava.antriksh96@gmail.com} \\
   \And
 Soumyashree Kar \\
  Center of Studies in Resources Engineering\\
  Indian Institute of Technology Bombay\\
  Mumbai, India 400076 \\
  \texttt{soumyakar@iitb.ac.in} \\
}

\begin{document}
\maketitle
\begin{abstract}
Graph neural networks (GNNs) offer a flexible framework for learning from scientific data with physical, biological, or functional associations. One promising domain is plant physiology, where observed responses result from several interacting processes that are difficult to isolate, even with human intervention. A key example is the A--Ci curve, which relates the net CO$_2$ assimilation rate ($A_\text{net}$) to leaf intercellular CO$_2$ concentration ($C_\text{i}$) and is also used to estimate photosynthetic parameters in biophysical models. However, accurate estimation requires accurate identification of the active biochemical limiting state at each curve point, which is a major source of uncertainty. Here, we express the limitation-state identification in A--Ci curves as a graph-based node classification problem. A graph representation of the A--Ci curve is created using distance-based k-nearest-neighbor (kNN) and auxiliary-signal-guided (ASG) connectivity. The methodology was evaluated against the conventional machine learning baselines, graph-based architectures, and an automated fitting-based benchmark. Results on a large synthetic dataset with known ground-truth limitation states show that graph-based models improve classification, especially near biochemical transition areas. The top-performing configuration, \textit{SEAGAN} (\textit{domain-Specific and Edge-Aware Graph Attention Network for Dynamic Plant Processes}), integrates process-aware node features, edge attributes, kNN connectivity, and graph attention with a weighted cross-entropy loss, obtaining an F1-score of 0.857 and accuracy of 0.882. The results suggest that analyzing A--Ci curves as graphs enables better identification of the biochemical limiting condition and reduces the uncertainty associated with both human and automated methods.
\end{abstract}

\keywords{Graph neural networks \and Photosynthetic parameters \and Node classification \and A--Ci curves \and Functional plant phenomics}

\section{Introduction}

The photosynthetic response of a species is commonly characterized using CO$_2$ response curves, also called A--Ci curves, which describe the relationship between net CO$_2$ assimilation rate ($A_\text{net}$) and intercellular CO$_2$ concentration ($C_\mathrm{i}$). This curve provides the basis for estimating key biochemical parameters of the Farquhar, von Caemmerer, and Berry (FvCB) model \citep{Farquhar1980}. The parameters derived from these curves, such as Rubisco carboxylation capacity ($V_\text{cmax}$), photosynthetic electron transport rate ($J$), triose phosphate utilization (TPU), mesophyll conductance ($g_\text{m}$), and day respiration ($R_\text{d}$), are widely used as mechanistically meaningful traits \citep{Sharkey2007}. They are used in studies ranging from leaf-level physiology \citep{Srivastava2024} to crop-canopy modeling \citep{Lochocki2022,Yulong2019}, and ecosystem-to-climate-scale modeling \citep{Bonan2014}. Accurate estimation of these parameters is central to understanding the plant physiological responses, comparing genotypic variation, and improving the process-based agricultural and ecological models.

The main challenge in estimating these parameters is correctly identifying the active biochemical limitation regime at each point along the A--Ci curve \citep{Gu2010,Moualeu2017}. The FvCB model operates fundamentally as a switch-type equation with discontinuous transitions, where $A_\text{net}$ at any given point is determined by the minimum of the rates limited by Rubisco activity, RuBP regeneration, or TPU \citep{Farquhar1980,Collatz1991,vonCaemmerer2000}. Assigning experimental data points to these distinct biochemical regimes is difficult because transitions between them are often unclear and highly ambiguous, further exacerbated by slight differences in the experiments, measurement errors, and intrinsic noise in portable gas-exchange data \citep{Long2003,Lei2025}.

Classical approaches to FvCB parameter estimation have often required manual assignment of biochemical limitation to each observation point in the A--Ci curves before fitting the corresponding model components \citep{Dubois2007,Sharkey2007,Moualeu2017}. Sharkey et al. \citep{Sharkey2007} proposed a rule-based heuristic approach for assigning these biochemical limitations, in which points below 200 ppm CO$_2$ are assigned to Rubisco limitation, points above 300 ppm CO$_2$ to RuBP-regeneration limitation, and points in the 200–300 ppm interval to an ambiguous transition zone. Such approaches are only useful for initial fitting and cannot be relied upon for all crop types. The standard gradient-based or quasi-Newton optimization algorithms struggled with the FvCB model because of its switch-type formulation, which creates discontinuities and irregular likelihood surfaces, resulting in the optimizer being trapped in local minima \citep{Lei2025,Lochocki2025}. Moreover, the need for manual or semi-manual intervention limits scalability, making these approaches difficult to use in high-throughput plant phenotyping and ecological studies, particularly when quantifying functional traits that extend beyond visually observable plant characteristics \citep{Furbank2011}.

To reduce these limitations, automated full-curve and exhaustive fitting approaches have been developed through software tools such as \textit{plantecophys} \citep{Duursma2015}, \textit{msuRACiFit} \citep{Gregory2021}, \textit{PhoTorch} \citep{Lei2025}, and \textit{PhotoGEA} \citep{Lochocki2025}. These methods aim to reduce the uncertainty introduced by manual intervention by simultaneously estimating the limiting states and the biochemical parameters during the optimization. These implementations also address some of the numerical difficulties of FvCB model fitting using more advanced computational strategies, including adaptive gradient optimization with biologically informed penalty terms in PhoTorch \citep{Lei2025} and derivative-free evolutionary search in \textit{PhotoGEA} \citep{Lochocki2025}. However, these advances do not resolve the primary identifiability issues. Additionally, when the limiting process is weakly expressed in the gas-exchange data, even state-of-the-art methods can yield unbounded and unreliable parameter estimates due to their heavy reliance on imposed penalties to force a fit. Thus, despite major progress in automated optimization, a methodological gap remains: the use of structural relationships between measurement points in the A--Ci curve to limit-state identification before photosynthetic parameter optimization.

Current fitting approaches primarily infer limiting regimes through global error minimization, even though local interactions among nearby or functionally related observations may hold crucial information for resolving ambiguous transition regions. The A--Ci curve is not just a collection of independent points, it comprises a structured trajectory in which nearby observations are connected through the underlying biochemistry. To leverage the structural interactions between points along the A--Ci curves, a new representation is necessary. Therefore, treating limitation-state identification as a node classification problem in a structured graph can separate it from global parameter optimization and reduce uncertainty in the subsequent FvCB parameter estimation.

Machine learning has already become an important tool in agricultural and plant-science applications, enabling crop-yield prediction from increasingly diverse data sources, including satellite and remote-sensing imagery \citep{You2017,Nevavuori2019}, UAV-based hyperspectral and high-throughput phenotyping data \citep{Riera2021,Li2022}, and genotype–environment or weather-related variables \citep{Khaki2019,Shook2021}. However, many conventional machine learning and deep learning approaches represent observations as independent samples, fixed-format feature vectors, or regular input arrays. Such representations are effective when the input structure is fixed, such as image-based tasks in computer vision. However, they are not suitable when the input array size cannot be fixed beforehand.

Graph neural networks (GNNs) overcome this limitation by allowing deep learning models to operate directly on irregular, non-grid, and relational data structures \citep{Zhou2020,Wu2021}. In GNNs, the graph structure is not fixed as in images or conventional neural networks (NN); it is constructed such that nodes, edges, and message-passing operations capture the governing interactions of the process(es) being modeled \citep{Franco2009,Bronstein2021}. This property has made GNNs highly effective when prediction depends not only on local features but also on interactions between connected components \citep{Battaglia2018,Sanchez2020}. Recent agricultural studies have also begun to use graph-based representations for crop-yield prediction by encoding geospatial and temporal relationships among samples \citep{Fan2022}, for smart-farming quality assessment \citep{Sajitha2023}, and for genotypic–topological modeling of field plots in which edges represent spatial and genotypic similarity \citep{Gupta2023}. Nevertheless, these applications mainly address crop-, field-, or system-level prediction tasks, while the use of graph-based learning for leaf-level physiological response curves remains largely unexplored. Adapting GNNs for such analysis requires defining how individual measurements are represented, how their relationships are encoded, and how to aggregate neighbors to reflect the underlying biological process. Thus, novelty in GNNs lies not only in architecture, but also in process-aware definitions of nodes, edges, and neighborhood aggregation \citep{Charilaos2014,Xu2018}.

Taken together, these limitations point to a specific methodological gap:
\begin{itemize}

    \item Current GNN applications in agriculture are mostly focused on crop-, field-, or system-level prediction, whereas leaf-level physiological response curves, such as A--Ci curves, remain underexplored.

    \item There is no graph-based framework for analyzing gas-exchange measurements.

    \item Uncertain identification of limiting processes due to their poor representation in a noisy A--Ci data.

    \item A methodological gap remains in detecting limiting regimes from structural connections in the A--Ci curve.

\end{itemize}

This study develops a domain-specific GNN framework for classifying the observation points along the A--Ci curves into distinct photosynthetic limitation states. The key contribution is to express the leaf-level CO$_2$ response analysis as a graph learning problem, where measurement points are represented as nodes, and their relationships are encoded as edges. A second contribution is the systematic evaluation of distinct GNN architectures and graph-construction strategies to identify the best-suited combination for A--Ci curve analysis. The frameworks evaluated include conventional machine learning (ML), deep learning (DL) \citep{LeCun2015}, graph convolutional networks (GCN) \citep{Thomas2017}, graph attention networks (GAT) \citep{Petar2018}, hierarchical Graph U-Net models \citep{Gao2019}, and an automated fitting-based benchmark \citep{Lei2025}. This assessment enables a direct comparison of how an increasingly expressive form of relational modeling performs relative to feature-based learning methods and the existing automated fitting-based regime identification approach. Graph Sample and Aggregation (GraphSAGE) was not included in the present comparison because it was primarily developed for inductive learning and scalable neighborhood sampling on large graphs \citep{William2017}. In contrast, the A--Ci graphs are small, containing only 8--15 nodes, and are fully observable for each curve.

To enable controlled evaluation, a large synthetic dataset of C3 A--Ci curves was generated using a wide range of biochemical parameter combinations with known ground-truth limiting states. This allows for a thorough model comparison while avoiding uncertainty introduced by manual labeling. 

The main contributions of this work are summarized as follows:
\begin{itemize}
    \item Formulation of A--Ci curve limitation-state as a node classification problem.
    \item Two graph-building algorithms are investigated: distance-based kNN and auxiliary-signal-guided (ASG) connectivity.
    \item Compare feature-based ML/DL baselines, GCN, GAT, Graph U-Net variations, and an automated fitting-based benchmark in a single assessment framework.
    \item All ground-truth limiting states are known, and evaluation is performed using a large synthetically generated dataset of A--Ci curves.
\end{itemize}

\section{Materials and methods}

\subsection{Synthetic generation of CO$_2$ response curves}
To train and evaluate the proposed GNN, we generated a large collection of synthetic C3 A--Ci curves.  In total, we simulated $N=10000$ curves with known ground-truth limitation regimes. Each synthetic curve was initially generated on a dense $C_\text{i}$ grid of 100 points from 20 (ppm) to 1000 (ppm), and then subsampled to retain 8--15 points to mimic the variable curve lengths typical of gas-exchange measurements.

To further improve realism, Gaussian noise was added to $A_\text{net}$. Across all simulated curves, the mean curve-wise standard deviation of the added error was $\approx 1.50~\si{\micro\mol\per\square\meter\per\second}$. When normalized by the standard deviation of the corresponding clean A--Ci curve, the added error represented $0.10$ times the curve-wise variability of $A_\text{net}$. This suggests that the perturbations are large enough to cause a measurable difference while not dominating the within-curve variation of $A_\text{net}$. A similar noise-addition approach has been used in previous studies to test the robustness of photosynthetic parameter-estimation models using synthetic gas-exchange data \citep{Zhou2019}. Furthermore, to ensure that the models learned the underlying physical meaning associated with the identification of the limiting process, rather than merely recognizing patterns in the ordering of limiting processes, the simulated A--Ci points were deliberately not arranged in order of increasing $C_\text{i}$.

Table~\ref{Tab:synth_ranges_params} summarizes the ranges of diffusive and biochemical parameters reported at $\SI{25}{\degreeCelsius}$, the usual reference temperature for studying photosynthetic parameters \citep{Sharkey2007}, and the leaf temperature ($T_\text{leaf}$) used in the simulation. To obtain broad coverage of the feasible parameter space while avoiding clustering that can occur under purely random sampling, we used Latin Hypercube Sampling (LHS) \citep{Mckay1979}. Temperature was treated as a curve-level constant and sampled uniformly within the specified bounds, whereas atmospheric pressure ($P_\mathrm{atm}=\SI{101.325}{\kilo\pascal}$) and O$_2$ partial pressure ($\mathrm{O}_2=\SI{210}{\milli\bar}$) were held fixed following standard A--Ci fitting assumptions \citep{Sharkey2007}.

\begin{table}[h]
\centering
\caption{Ranges of the biochemical and diffusive parameters used to generate synthetic C3 CO$_2$ response curves. For variable definitions, see Table~S1} \label{Tab:synth_ranges_params}
\vspace{0.1cm}
\begin{tabular}{p{5cm}|c}
\hline
\textbf{Parameter} & \textbf{Range} \\
\hline
$V_\text{cmax,25} \quad (\si{\micro\mol\per\square\meter\per\second})$ & 20--150 \\
$\mathrm{TPU}_\text{25} \quad (\si{\micro\mol\per\square\meter\per\second})$ & 4--25 \\
$R_\text{d,25} \quad (\si{\micro\mol\per\square\meter\per\second})$ & 0.5--5 \\
$g_\text{m,25} \quad (\si{\mol\per\square\meter\per\second})$ & 0.05--0.8 \\
$J_\text{max,25}/V_{cmax,25} \quad (-)$ & 1.5--2.5 \\
$T_\text{leaf} \quad (\SI{}{\degreeCelsius})$ & 25--35 \\
\hline
\end{tabular}
\end{table}

\subsubsection{C3 photosynthesis}
\begin{figure}[!t]
\centering
\includegraphics[width=0.45
\textwidth]{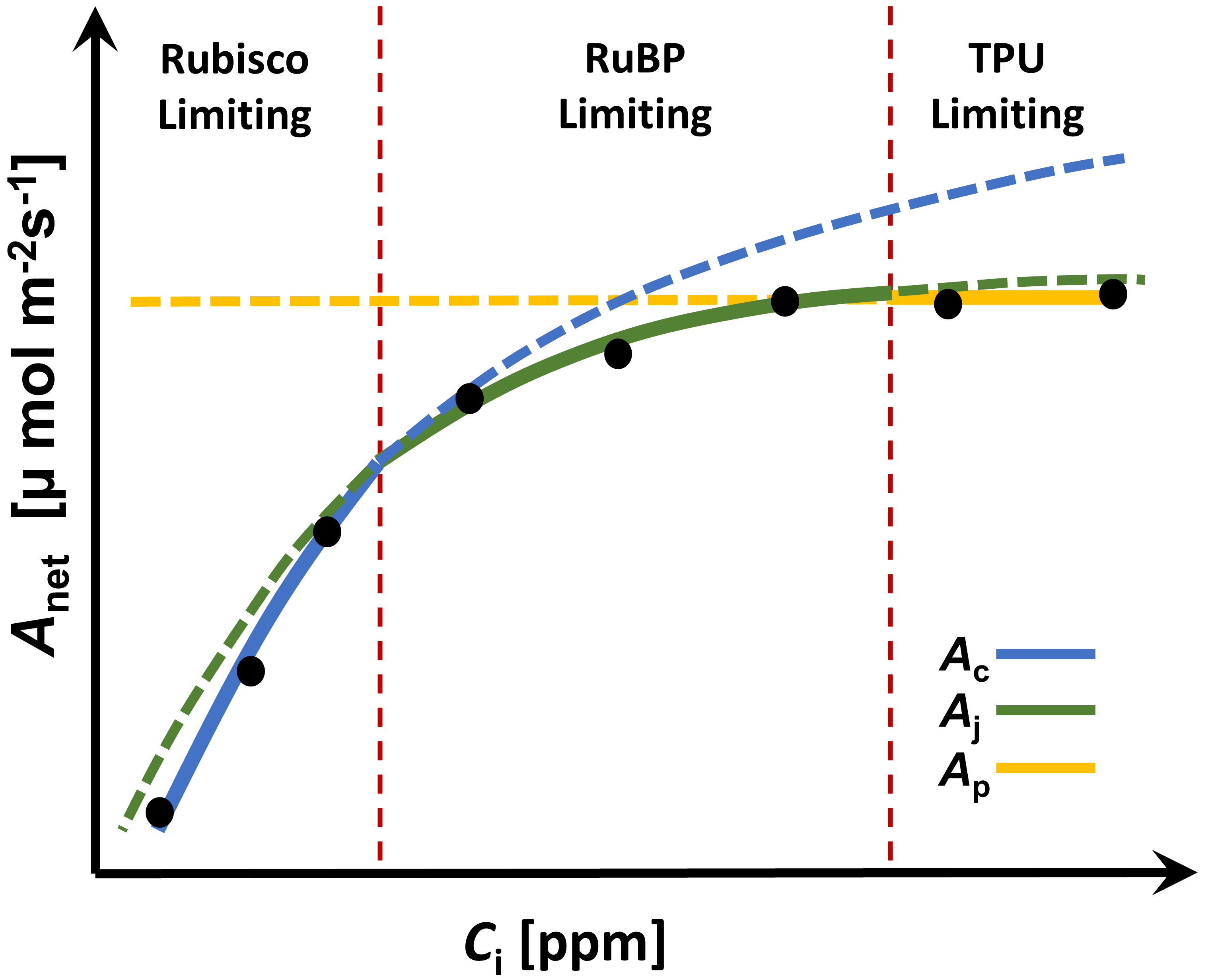}
\caption{Schematic C3 A--Ci curve showing limitation regimes and transition points. Black circles are measured net CO$_2$ assimilation ($A_\text{net}$) at discrete intercellular CO$_2$ ($C_\mathrm{i}$) levels. Solid lines indicate where each limiting process controls $A_\text{net}$. Colors represent candidate biochemical limitation rates: Rubisco-limited ($A_\text{c}$), RuBP-regeneration/electron-transport-limited ($A_\text{j}$), and TPU-limited ($A_\text{p}$).} \label{Fig:C3ACi}
\end{figure}

To analyze A--Ci curves in C3 plants, we use the biochemical photosynthesis framework, in which distinct limiting processes influence the net CO$_2$ assimilation rate ($A_\text{net}$) \citep{Farquhar1980}. The primary limitation changes from a Rubisco (carboxylation) ($A_\text{c}$) limitation to a RuBP-regeneration (electron-transport) ($A_\text{j}$) limitation with increasing $C_\text{i}$, and in certain cases to a triose-phosphate utilization (TPU) ($A_\text{p}$) limitation \citep{Farquhar1980,vonCaemmerer2000}. These regime shifts are the mechanistic basis for the characteristic change in slope and curvature observed along the A--Ci curve (Fig. \ref{Fig:C3ACi}). $A_\text{net}$ is therefore written as:

\begin{equation}
A_\text{net} = \min\!\left\{A_\text{c},\;A_\text{j},\;A_\text{p}\right\} \label{Eq:AnetC3}
\end{equation}

Each limiting process is modeled based on the following equations:

\begin{equation}
\begin{aligned}
A_\text{c} = W_\text{c} - R_\text{d} & \quad ; W_\text{c} = V_\text{cmax} \, \frac{C_\text{c} - \Gamma^\ast}{\displaystyle C_\text{c} + K_\text{c}\left(1 + \frac{O_\text{c}}{K_\text{o}}\right)} \\
A_\text{j} = W_\text{j} - R_\text{d} & \quad ; W_\text{j} = \frac{J}{4}\ \frac{C_\text{c} - \Gamma^\ast}{\displaystyle C_\text{c} + 2\Gamma^\ast}\\
A_\text{p} = W_\text{p} - R_\text{d} & \quad ; W_\text{p} = 3\ \mathrm{TPU}
\end{aligned} \label{Eq:AnetLimitation}
\end{equation}

where $W_\text{c}$, $W_\text{j}$, and $W_\text{p}$ are the gross CO$_2$ assimilation rate ($\si{\micro\mol\per\square\meter\per\second}$) based on Rubisco-limitation, RuBP regeneration, and triose-phosphate utilization, respectively. $R_\text{d}$ ($\si{\micro\mol\per\square\meter\per\second}$) is the dark respiration rate, $K_\text{c}$ ($\si{\micro\bar}$), and $K_\text{o}$ ($\si{\milli\bar}$) are the Michaelis–Menten constants of Rubisco for CO$_2$ and O$_2$, respectively. $O_\text{c}$ ($\si{\milli\bar}$) is the O$_2$ concentration in the chloroplast, $J$ ($\si{\micro\mol\per\square\meter\per\second}$) is the electron transport rate, and $\Gamma^\ast$ (ppm) is the CO$_2$ compensation point in the absence of mitochondrial respiration. $C_\text{c}$ (ppm) is the chloroplast CO$_2$ concentration. If mesophyll conductance ($g_\text{m}$) ($\si{\mol\per\square\meter\per\second}$) is neglected, $C_\text{c}$ is approximated as $C_\text{i}$ (ppm). When $g_\text{m}$ is included, $C_\text{c}$ is calculated as:  

\begin{equation}
C_\text{c} = C_\text{i} - \frac{A_\text{net}}{g_\text{m}} \label{Eq:Cc}
\end{equation}

The temperature-response functions used to simulate biochemical parameters at different leaf temperatures (Table~\ref{Tab:synth_ranges_params}), along with the functions for \(K_\text{c}\), \(K_\text{o}\), and \(\Gamma^\ast\), were adopted from Sharkey et al.~\citep{Sharkey2007}. The parameter values for these functions were taken from Harley et al. \citep{Harley1992} and Bernacchi et al. \citep{Bernacchi2001,Bernacchi2002,Bernacchi2003}.

\subsection{Auxiliary Signal Construction} \label{Sec:AuxSignal}
To assist classification of nodes in the A--Ci curve, we incorporate two supplementary features that capture Rubisco- and electron-transport-related patterns, respectively, providing additional information beyond $C_\text{i}$ and $A_\text{net}$ values. For each measurement point $j$, the auxiliary response factors were defined as:

\begin{equation}
\begin{aligned}
\phi_c(j) &= \frac{C_\text{i}(j)-\Gamma^*}{\displaystyle C_\text{i}(j) + K_\text{c}\left(1+\frac{O_\text{c}}{K_\text{o}}\right)}, \\
\phi_j(j) &= \frac{1}{4}\frac{C_\text{i}(j)-\Gamma^*}{\displaystyle C_\text{i}(j) + 2\Gamma^*},
\end{aligned}
\end{equation}

where $\phi_c$ and $\phi_j$ are dimensionless $C_\text{i}$-based response terms derived from the Rubisco-limited and electron-transport-limited gross assimilation equations, with $C_\text{i}$ in place of $C_\text{c}$ \Eqref{Eq:AnetLimitation}. These values were then used to normalize the observed $A_\text{net}$ values.

\begin{equation}
\begin{aligned}
s_{Ac}(j) &= \frac{A_{\text{net}}(j)}{\phi_c(j)}, \\
s_{Aj}(j) &= \frac{A_{\text{net}}(j)}{\phi_j(j)}.
\end{aligned}
\end{equation}

\begin{figure*}[!t]
    \centering
    \includegraphics[width=\textwidth]{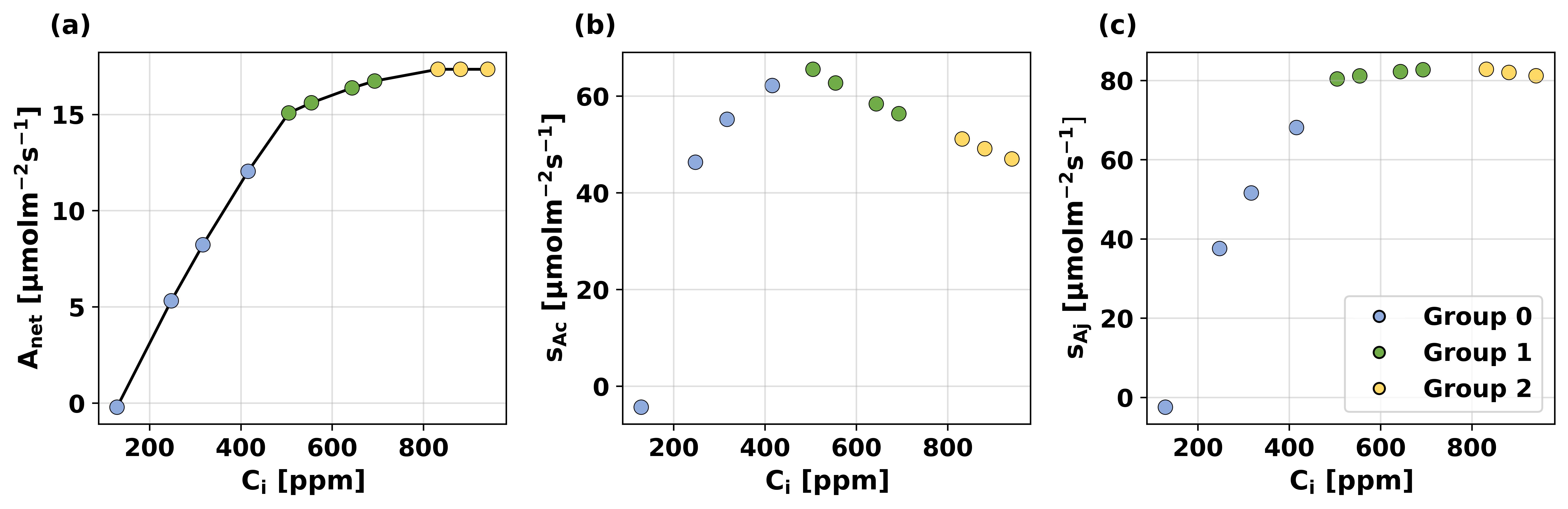}
    \caption{Auxiliary signals from A--Ci response curve. (a) Example A--Ci curve with node labels (Groups 0--2). (b--c) Corresponding auxiliary signals $s_{Ac}$ and $s_{Aj}$ plotted against $C_\mathrm{i}$, showing how the features separate the three classes.}
    \label{Fig:EdgeDetection}
\end{figure*}

Normalized $A_\text{net}$ measurements, $s_{Ac}$, and $s_{Aj}$ were supplementary diagnostic variables. These show systematic changes over the A--Ci response curve and hence support the identification of various limiting regimes (Fig.~\ref{Fig:EdgeDetection}). $s_{Ac}$ only increases in the Rubisco-limited zone, while $s_{Aj}$ increases until the TPU-limited region and then decreases (\figpanels{Fig:EdgeDetection}{b-c}). However, the noise in the observed A--Ci responses does not allow a robust differentiation along the A--Ci curve based only on these patterns. Instead, the signals are employed as additional input features, and for graph-based models, are further used to establish connections between measurement points (Section~\ref{Sec:GroupAssignment}).

\subsection{Graph construction} \label{Sec:GraphConstruction}
\subsubsection{Node representation}
Each synthetic A--Ci curve was pre-processed and then represented as an individual graph. Each measurement point on the curve was considered as a node. Therefore, the graph corresponding to a curve with m sampled A--Ci points has $m$ nodes, each indexed by $j=1,\dots,m$. The number of directed edges in the graph is denoted by $n$.

The feature vector assigned to node $j$ contains two directly observed gas-exchange quantities and two auxiliary signals derived from the local physiological response of the curve. The observed quantities are $C_\mathrm{i}(j)$ and $A_{\mathrm{net}}(j)$. The auxiliary quantities $s_{Ac}(j)$ and $s_{Aj}(j)$ are designed to provide additional information on the Rubisco-limited and RuBP-regeneration-limited portions of the A--Ci response, respectively. The node feature vector is therefore defined as

\begin{equation}
\mathbf{x}_j =
\begin{bmatrix}
C_\mathrm{i}(j) \\
A_{\mathrm{net}}(j) \\
s_{Ac}(j) \\
s_{Aj}(j)
\end{bmatrix},
\qquad
\mathbf{x}_j \in \mathbb{R}^{4}.
\label{Eq:NodeLabel}
\end{equation}

For each graph, the feature vectors of all nodes are concatenated to get the entire node-feature matrix. This structure allows each node to keep its pointwise A--Ci information, and the graph edges provided in the next sections indicate how information is transferred across connected measurement points during message passing.

\subsubsection{Group assignment} \label{Sec:GroupAssignment}

The auxiliary signals defined in Section~\ref{Sec:AuxSignal} were used to divide each A--Ci curve into contiguous response regions for ASG graph construction. Because the sampled points of an A--Ci curve were not necessarily stored in increasing order of $C_\mathrm{i}$, group assignment was first performed on the $C_\mathrm{i}$-sorted curve and then mapped back to the original node order.

Let \(\pi(r)\) denote the original node index of the point occupying rank \(r\) after sorting the curve in ascending order of \(C_\mathrm{i}\), where \(r=1,\dots,m\). The sorted quantities are then written as
\begin{equation}
\tilde{C}_\mathrm{i}(r) = C_\mathrm{i}(\pi(r)), \qquad
\tilde{A}_{\mathrm{net}}(r) = A_{\mathrm{net}}(\pi(r)),
\end{equation}
with corresponding sorted auxiliary signals
\begin{equation}
\tilde{s}_{Ac}(r) = s_{Ac}(\pi(r)), \qquad
\tilde{s}_{Aj}(r) = s_{Aj}(\pi(r)).
\end{equation}

Peak detection was applied to the sorted auxiliary-signal sequences to identify two characteristic transition indices:
\begin{equation}
p_c = \mathcal{P}\!\left(\tilde{s}_{Ac}\right), \qquad
p_j = \mathcal{P}\!\left(\tilde{s}_{Aj}\right),
\end{equation}
where \(\mathcal{P}(\cdot)\) denotes the peak-detection operator. In this study, \(\mathcal{P}(\cdot)\) identifies the dominant local peak in each sorted auxiliary-signal sequence after pre-processing.

Using these peak indices, the sorted curve was divided into three contiguous rank-index groups:
\begin{equation}
\begin{aligned}
\tilde{\mathcal{G}}_0 &= \{\, r : r \le p_c \,\},\\
\tilde{\mathcal{G}}_1 &= \{\, r : p_c < r \le p_j \,\},\\
\tilde{\mathcal{G}}_2 &= \{\, r : r > p_j \,\}.
\end{aligned}
\label{Eq:GroupAssignSorted}
\end{equation}

Because graph construction is performed using the original node indices, these sorted groups were mapped back to the original node order as
\begin{equation}
\mathcal{G}_q =
\{\, \pi(r) : r \in \tilde{\mathcal{G}}_q \,\},
\qquad q \in \{0,1,2\}.
\label{Eq:GroupAssignOriginal}
\end{equation}

The resulting auxiliary-signal groups were used to define graph connectivity and edge construction in the ASG graph (Fig.~\ref{Fig:GraphConnectivity}). They were not used as ground-truth limitation labels. Instead, they offer a systematic, data-guided segmentation of the A--Ci curve into sections with different response behavior.

\subsubsection{Edge representation for message passing} \label{Sec:EdgeAttr}

Edges define how information is exchanged between measurement points within the same A--Ci curve. In this study, we consider two graph connectivity schemes: a distance-based $k$-nearest neighbors (kNN) graph and an auxiliary-signal-guided (ASG) graph (Section~\ref{Sec:GroupAssignment}). In both schemes, the edge attributes are defined using the auxiliary signals. Hereafter, the two graph settings are referred to by their construction principles rather than by their edge densities: the distance-based k-nearest-neighbor (kNN) graph and the auxiliary-signal-guided (ASG) graph. \\

\noindent\textit{Distance-based kNN connection: }
In this setup, each node is connected to its $k = 4$ nearest neighbors, excluding itself, based on the Euclidean distance in the A--Ci curve space. The value $k = 4$ was selected based on a sensitivity analysis of both GCN- and GAT-based kNN graph models (Fig.~S1). For the GCN model, the best performance was observed at $k = 4$, while in the GAT model, limited gains were observed for $k>4$ (Fig.~S1). Hence, $k = 4$ was used to retain the A--Ci curve structure while ensuring a consistent graph construction across the model comparisons. Let $\mathcal{N}_k(j)$ denote the set of the 4 nearest neighbors of node $j$. Directed edges are created from node $j$ to all nodes in $\mathcal{N}_k(j)$, and the graph is then made bidirectional by adding the corresponding reverse edges. Thus, the kNN edge set is written as

\begin{equation}
\mathcal{E}_{\mathrm{kNN}}
=
\{(u,v): v \in \mathcal{N}_k(u)\}
\;\cup\;
\{(v,u): v \in \mathcal{N}_k(u)\}.
\end{equation}

\begin{figure}[!t]
\centering
\includegraphics[width=0.6\textwidth]{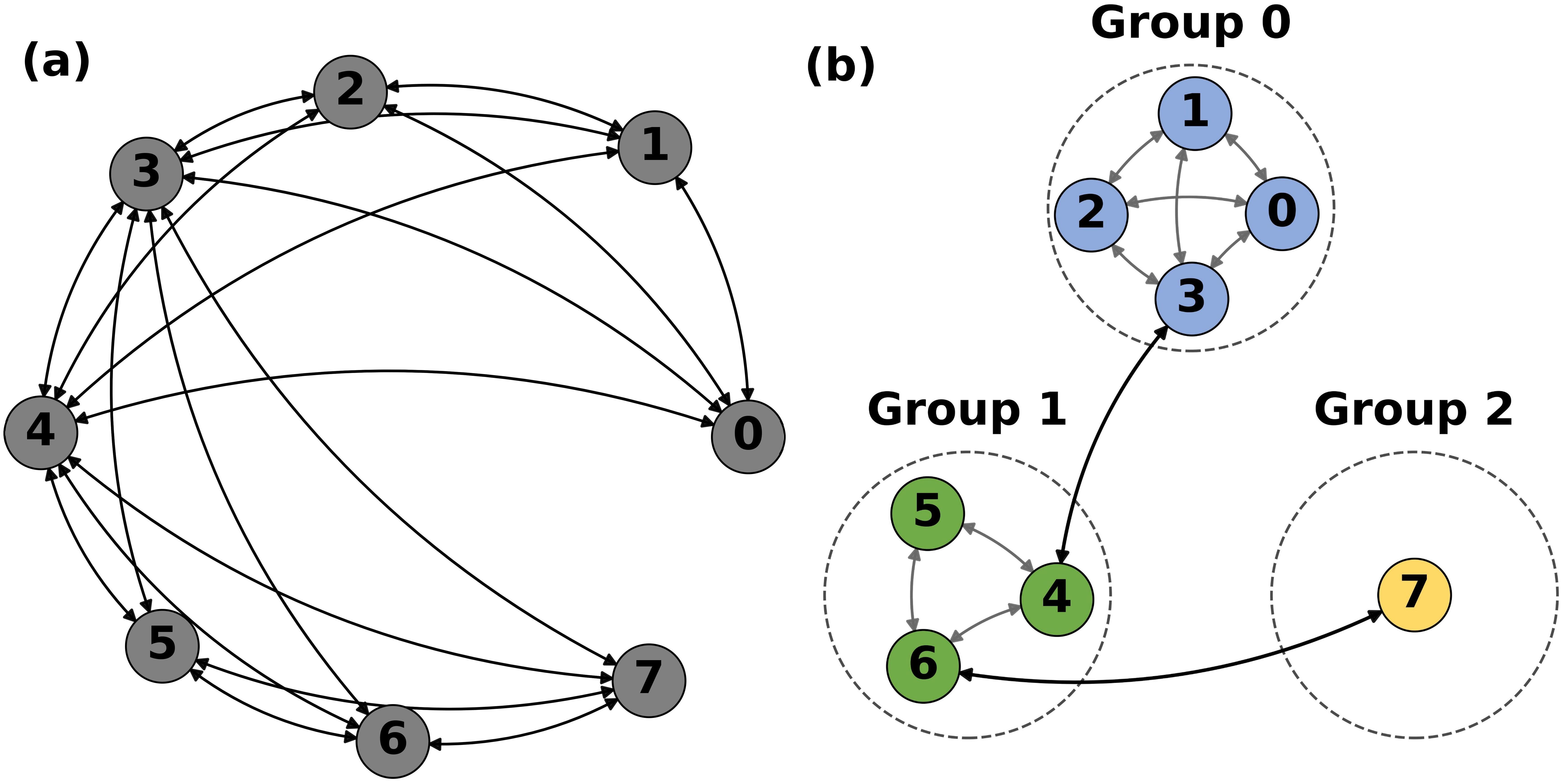}
\caption{Graph construction strategies for the A--Ci response curve. (a) Distance-based kNN connection: A proximity-based graph where each node is bidirectionally connected to its $k$=4 nearest neighbors using Euclidean distance. (b) Auxiliary-Signal-Guided (ASG) connection: An informed graph where nodes are fully connected within predefined functional groups (Groups 0--2). Boundary nodes provide bridging links between groups to maintain global continuity.} \label{Fig:GraphConnectivity} 
\end{figure}

As a result, each node is guaranteed to connect to at least four nearby neighbors, while the total number of incident edges may exceed four because a node can also receive reverse connections from other nodes for which it lies among their nearest neighbors. This distance-based construction allows richer local message passing along the curve (\figpanels{Fig:GraphConnectivity}{a}). For every directed edge $(u,v)\in\mathcal{E}_{\mathrm{kNN}}$, the edge attribute vector is defined below.\\

\noindent\textit{Auxiliary-Signal-Guided (ASG) connection: }
For the ASG setting, edges are constructed using the group assignment defined in Section~\ref{Sec:GroupAssignment}. Nodes within a group are fully connected, and connections between boundary nodes of neighboring groups are added to maintain continuity across the graph (\figpanels{Fig:GraphConnectivity}{b}). All connections are bidirectional in nature. This produces a structured graph in which connectivity follows the biologically guided segmentation of the response trajectory. Interestingly, ASG connectivity was not uniformly sparser or denser than kNN connectivity. ASG tends to produce lower densities for shorter curves but becomes comparable or even denser than kNN for curves with 12 or more points. This shows that the two graph-construction methods differ not only in their physiological basis but also in the way their connectivity scales. As in the kNN graph, edge attributes are assigned to each edge following the definition in the section below.

\begin{equation}
\begin{aligned}
\mathcal{E}_{\mathrm{ASG}}
&= \bigcup_{g=0}^{2}
\{(u,v): u,v \in \mathcal{G}_g,\; u \neq v\}
\;\cup\;
\mathcal{E}_{\mathrm{boundary}}, \\
\mathcal{E}_{\mathrm{boundary}}
&= \{(u_{01},v_{01}), (v_{01},u_{01}), (u_{12},v_{12}), (v_{12},u_{12})\},
\end{aligned}
\end{equation}
where
\begin{equation}
\begin{aligned}
u_{01} = \max(\mathcal{G}_0), \qquad
v_{01} = \min(\mathcal{G}_1), \\
u_{12} = \max(\mathcal{G}_1), \qquad
v_{12} = \min(\mathcal{G}_2).
\end{aligned}
\end{equation}

\noindent\textit{Edge attributes: } 
Edges encode relations between the points of the A--Ci curve. In the graph, for every pair of linked nodes \(\{u,v\}\), two directed edges \((u,v)\) and \((v,u)\) are included. For each ordered edge \((u,v)\in\mathcal{E}\), the edge attribute vector is given by differences in the auxiliary node-level signals:

\begin{equation}
\mathbf{e}_{uv} = 
\begin{bmatrix}
e_{c,uv} \\
e_{j,uv}
\end{bmatrix} =
\begin{bmatrix}
s_{Ac}(u) - s_{Ac}(v) \\
s_{Aj}(u) - s_{Aj}(v)
\end{bmatrix} \in \mathbb{R}^{2}
\label{Eq:Edgelabel}
\end{equation}

Since both directions are included, the reverse edge has the corresponding opposite-valued attribute,
\(\mathbf{e}_{vu}=-\mathbf{e}_{uv}\). These edge features encode local changes in the regime-sensitive auxiliary signals between connected measurement points, providing the graph attention mechanism with relational information that complements the node features while preserving bidirectional message passing.

\subsection{Methodological framework} 

\begin{figure*}[!t]
    \centering
    \includegraphics[width=\textwidth]{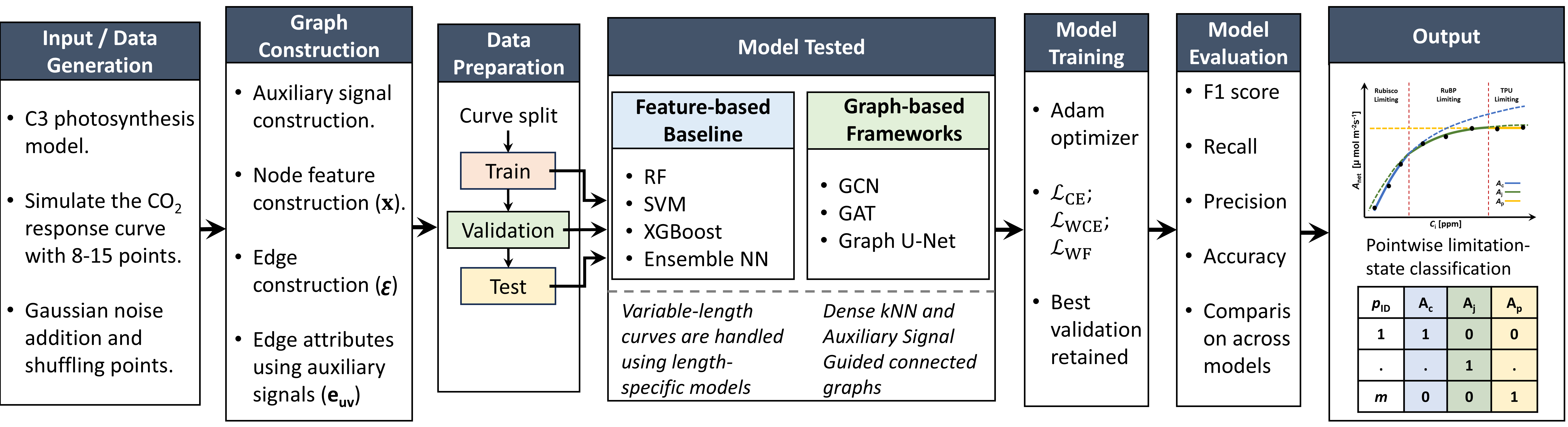}
    \caption{Workflow for automated node-wise classification of photosynthetic limitation states from synthetic CO$_2$ response curves. The framework includes curve generation, graph construction, data preparation, development of feature-based and graph-based models, model training under multiple loss functions, and comparative evaluation using standard classification metrics, leading to automated segmentation of the \(A_c\)-, \(A_j\)-, and \(A_p\)-limited regions.} \label{Fig:Methodology_Overview}
\end{figure*}

\subsubsection{Methodology overview}
The overall methodology is summarized in Fig.~\ref{Fig:Methodology_Overview}. The workflow begins with the synthetic generation of C3 A--Ci curves. The curves are then subsampled, and noise is added to emulate actual gas-exchange measurements. Node-level features were constructed and used by all models, while graph-based models additionally used either kNN or ASG edge construction and edge attributes to represent relationships among measurement points.

The A--Ci curves and their graphical representations were then partitioned into training, validation, and test subsets. To benchmark the proposed framework against non-graphic techniques, feature-based baseline models were used. Random Forest (RF) \citep{Breiman2001}, Support Vector Machine (SVM) \citep{Cortes1995}, XGBoost \citep{Chen2016}, and an ensemble feed-forward neural network \citep{LeCun2015} were trained using only the node features \Eqref{Eq:NodeLabel}. Since these ML baselines do not support variable-length inputs, individual models were trained for each data subset based on the curve lengths.

In parallel, graph-based representations of the same curves were used to train the GCN, GAT, and Graph U-Net frameworks. From neighborhood aggregation with fixed graph connectivity to attention-based weights and hierarchical graph pooling, these models progressively raise the learning complexity. All models were trained using the Adam optimizer in combination with different loss formulations. For all model trainings, the checkpoint with the highest macro-F1-score on the validation set was retained for final testing.

\subsubsection{ML baseline}
\noindent{\textit{Standard ML models:}}
As feature-based baselines, we evaluated three standard supervised learning models: RF, SVM, and XGBoost. For all three cases, the classification was done only with the node-level feature representation \(\mathbf{x}_j\) \Eqref{Eq:NodeLabel}. These models did not consider each A--Ci point as independent. Instead, the entire curve was fed into the models at once by concatenating the node feature vectors \(\mathbf{x}_j\) of all points into a single, flattened curve-level input vector of size $4m$. The corresponding output is a vector of size $m$ containing the limitation-state labels. These models therefore provide a reference for how well the photosynthetic limitation state can be inferred from pointwise feature values alone. Because the dataset contains A--Ci curves ranging from 8 to 15 points, dedicated models were trained on a subset, based on curve length, and the appropriate model was selected during inference. \\

\noindent{\textit{Ensemble NN model:}}
To provide a non-graph deep learning baseline, we used an ensemble of feed-forward neural networks, with one network trained for each curve length \(m \in \{8,\dots,15\}\). For a curve with $m$ sampled points, the input was arranged as a feature matrix of size $[m,4]$, containing the pointwise variables \([C_\mathrm{i},\; A_{\mathrm{net}},\;s_{Ac},\;s_{Aj}]\), and then flattened into a vector of size $[4m,1]$ before being passed to the corresponding network. Each network consisted of fully connected hidden layers followed by a final linear output layer producing $m \times 3$ logits, corresponding to the three limitation classes predicted for all points in the curve. Unlike standard ML models, the NN baseline can learn complex nonlinear interactions across the full ordered feature space of a curve; however, it still does not leverage graph connectivity or structure.

\begin{table*}[t]
\centering
\caption{Summary of model scenarios used in the comparative evaluation. A tick indicates the presence of a component, and a cross indicates its absence. Here, kNN denotes distance-based k-nearest-neighbor connectivity, and ASG denotes auxiliary-signal-guided connectivity.}
\label{Tab:scenario_summary}
\begin{tabular}{p{0.2cm} p{2.8cm} p{0.5cm} l p{0.5cm}p{0.5cm} p{0.6cm}p{1.15cm}p{0.9cm} l}
\hline
\multirow{2}{*}{\textbf{ID}} 
& \multirow{2}{=}{\textbf{Model category}} 
& \multicolumn{2}{c}{\textbf{Model}}
& \multicolumn{2}{c}{\textbf{Connections}}
& \multicolumn{3}{c}{\textbf{Graph}} 
& \multirow{2}{*}{\textbf{Loss}} \\
\cline{3-9}
& & \textbf{Code} & \textbf{Name} & \textbf{ASG} & \textbf{kNN} & \textbf{Conv.} & \textbf{Attention} & \textbf{U-Net} &  \\
\hline

1 & DL baseline & 1-1 & NN 
& \xmark & \xmark 
& \xmark & \xmark & \xmark
& $\mathcal{L}_{\mathrm{CE}}$ \\
\hline

2 & GNN baseline & 2-1 & GCN-kNN-$\mathcal{L}_\mathrm{CE}$ 
 &\xmark & \cmark 
& \cmark & \xmark & \xmark
& $\mathcal{L}_{\mathrm{CE}}$ \\
\hline

\multirow{3}{*}{3}
& \multirow{3}{=}{Graph U-Net + kNN}
& 3-1 & GCN-U-Net-kNN-$\mathcal{L}_\mathrm{CE}$ 
& \xmark & \cmark 
& \cmark & \xmark & \cmark
& $\mathcal{L}_{\mathrm{CE}}$ \\
&  & 3-2 & GCN-U-Net-kNN-$\mathcal{L}_\mathrm{WCE}$ 
& \xmark & \cmark 
& \cmark & \xmark & \cmark
& $\mathcal{L}_{\mathrm{WCE}}$ \\
&  & 3-3 & GCN-U-Net-kNN-$\mathcal{L}_\mathrm{WF}$ 
& \xmark & \cmark 
& \cmark & \xmark & \cmark
& $\mathcal{L}_{\mathrm{WF}}$ \\
\hline

\multirow{3}{*}{4}
& \multirow{3}{=}{Graph U-Net + ASG}
& 4-1 & GCN-U-Net-ASG-$\mathcal{L}_\mathrm{CE}$ 
& \cmark & \xmark 
& \cmark & \xmark & \cmark
& $\mathcal{L}_{\mathrm{CE}}$ \\
&  & 4-2 & GCN-U-Net-ASG-$\mathcal{L}_\mathrm{WCE}$ 
& \cmark & \xmark 
& \cmark & \xmark & \cmark
& $\mathcal{L}_{\mathrm{WCE}}$ \\
&  & 4-3 & GCN-U-Net-ASG-$\mathcal{L}_\mathrm{WF}$ 
& \cmark & \xmark 
& \cmark & \xmark & \cmark
& $\mathcal{L}_{\mathrm{WF}}$ \\
\hline

\multirow{3}{*}{5}
& \multirow{3}{=}{Graph + Attention + kNN}
& 5-1 & GAT-kNN-$\mathcal{L}_\mathrm{CE}$ 
& \xmark & \cmark 
& \xmark & \cmark & \xmark
& $\mathcal{L}_{\mathrm{CE}}$ \\
&  & 5-2 & GAT-kNN-$\mathcal{L}_\mathrm{WCE}$ 
& \xmark & \cmark 
& \xmark & \cmark & \xmark
& $\mathcal{L}_{\mathrm{WCE}}$ \\
&  & 5-3 & GAT-kNN-$\mathcal{L}_\mathrm{WF}$ 
& \xmark & \cmark 
& \xmark & \cmark & \xmark
& $\mathcal{L}_{\mathrm{WF}}$ \\
\hline

\multirow{3}{*}{6}
& \multirow{3}{=}{Graph + Attention + ASG}
& 6-1 & GAT-ASG-$\mathcal{L}_\mathrm{CE}$ 
& \cmark & \xmark 
& \xmark & \cmark & \xmark
& $\mathcal{L}_{\mathrm{CE}}$ \\
&  & 6-2 & GAT-ASG-$\mathcal{L}_\mathrm{WCE}$ 
& \cmark & \xmark 
& \xmark & \cmark & \xmark
& $\mathcal{L}_{\mathrm{WCE}}$ \\
&  & 6-3 & GAT-ASG-$\mathcal{L}_\mathrm{WF}$ 
& \cmark & \xmark 
& \xmark & \cmark & \xmark
& $\mathcal{L}_{\mathrm{WF}}$ \\
\hline

\multirow{3}{*}{7}
& \multirow{3}{=}{Graph U-Net + Attention + kNN}
& 7-1 & GAT-U-Net-kNN-$\mathcal{L}_\mathrm{CE}$ 
& \xmark & \cmark 
& \xmark & \cmark & \cmark
& $\mathcal{L}_{\mathrm{CE}}$ \\
&  & 7-2 & GAT-U-Net-kNN-$\mathcal{L}_\mathrm{WCE}$ 
& \xmark & \cmark 
& \xmark & \cmark & \cmark
& $\mathcal{L}_{\mathrm{WCE}}$ \\
&  & 7-3 & GAT-U-Net-kNN-$\mathcal{L}_\mathrm{WF}$ 
& \xmark & \cmark 
& \xmark & \cmark & \cmark
& $\mathcal{L}_{\mathrm{WF}}$ \\
\hline

\multirow{3}{*}{8}
& \multirow{3}{=}{Graph U-Net + Attention + ASG}
& 8-1 & GAT-U-Net-ASG-$\mathcal{L}_\mathrm{CE}$ 
& \cmark & \xmark 
& \xmark & \cmark & \cmark
& $\mathcal{L}_{\mathrm{CE}}$ \\
&  & 8-2 & GAT-U-Net-ASG-$\mathcal{L}_\mathrm{WCE}$ 
& \cmark & \xmark 
& \xmark & \cmark & \cmark
& $\mathcal{L}_{\mathrm{WCE}}$ \\
&  & 8-3 & GAT-U-Net-ASG-$\mathcal{L}_\mathrm{WF}$ 
& \cmark & \xmark 
& \xmark & \cmark & \cmark
& $\mathcal{L}_{\mathrm{WF}}$ \\
\hline
\end{tabular}
\end{table*}

\subsubsection{Graph neural network baseline (GCN)}

\begin{figure*}
\centering
    \includegraphics[width=1.05\textwidth]{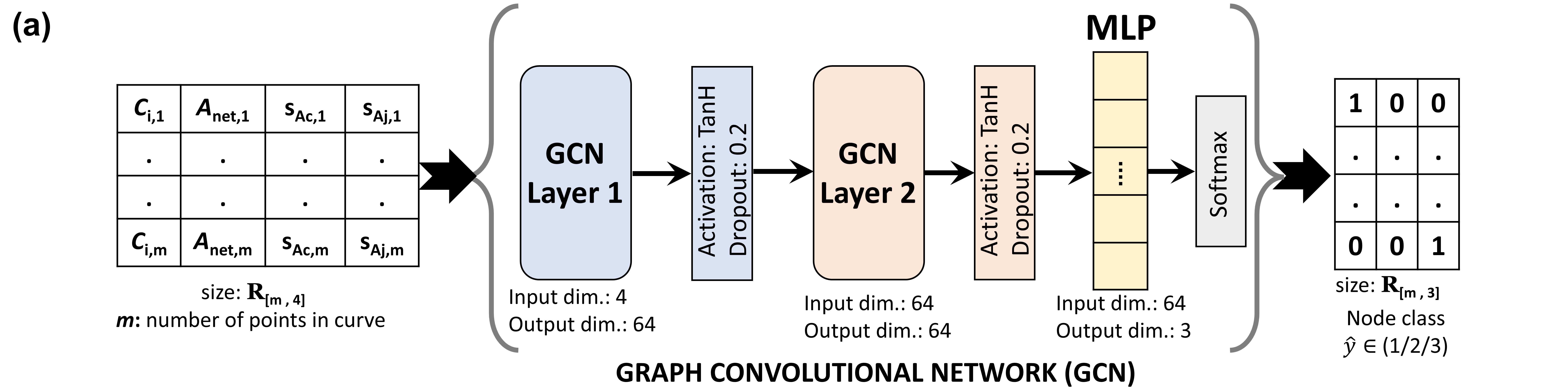}
    \vspace{1em}
    \includegraphics[width=1.05\textwidth]{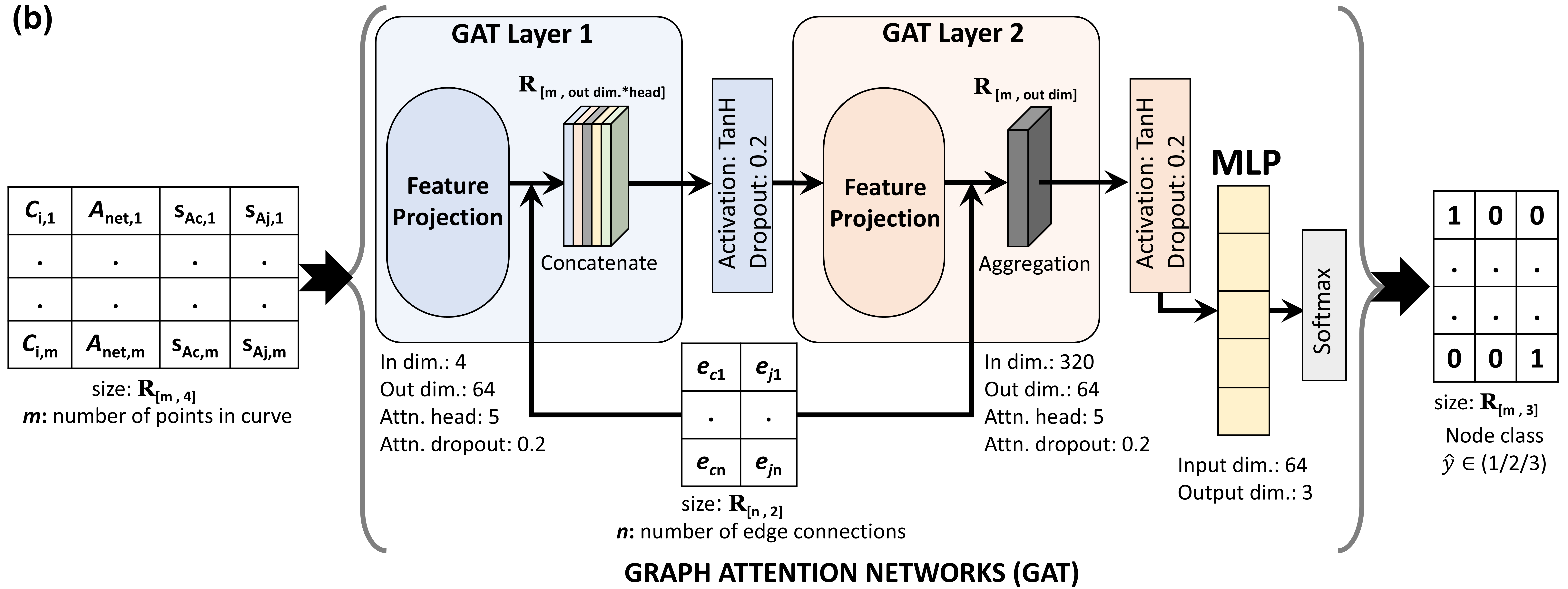}
    \vspace{1em}
    \includegraphics[width=1.05\textwidth]{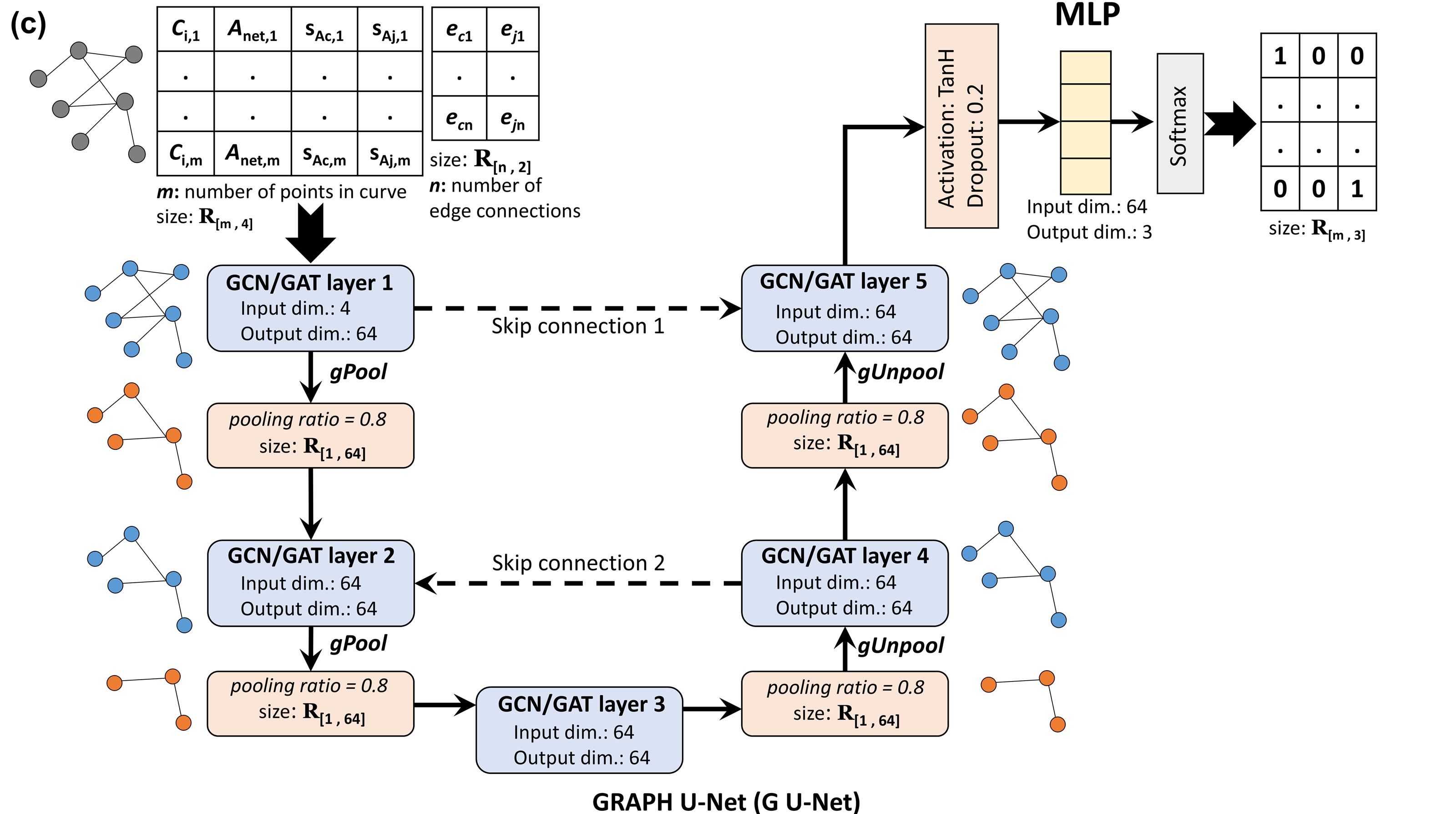}
\caption{Graph-based model frameworks for limitation-state classification. Workflow of the graph-based model frameworks used for node-wise classification of photosynthetic limitation states along CO$_2$ response curves: (a) GCN baseline, (b) GAT with edge-aware attention, and (c) Graph U-Net with hierarchical pooling and unpooling.}
\label{Fig:Graph_Model_Frameworks}
\end{figure*}

The graph convolutional network (GCN) was considered as the GNN baseline (\figpanels{Fig:Graph_Model_Frameworks}{a}). Each curve was represented as a graph \(G = (\mathcal{V}, \mathcal{E})\) with the nodes corresponding to the A--Ci curve points and the node features were defined as in Eq.~\ref{Eq:NodeLabel}. The graph connectivity was constructed using the bidirectional kNN scheme described in Section~\ref{Sec:EdgeAttr}. Thus, each node exchanged information with at least 4 of its nearest neighbors.

The GCN model consists of stacked graph convolution layers followed by batch normalization, \(tanh\) activation, and dropout (\(p = 0.2\)). Given that \(\text{h}_j^{(\ell)}\) is the hidden representation of node $j$ at layer \(\ell\), then \(\text{H}^{(\ell)}\) denotes the stacked matrix formed by all nodes at that layer. The graph convolution update can therefore be written in compact form as:

\begin{equation}
\text{H}^{(\ell+1)} =
\sigma\!\left(\hat{\text{A}}\,\text{H}^{(\ell)}\text{W}^{(\ell)}\right)
\end{equation}

where \(\hat{\text{A}}\) is the normalized graph connectivity matrix, \(\mathbf{W}^{(\ell)}\) is the learnable weight matrix, and \(\sigma(\cdot)\) denotes the nonlinear transformation. After the final graph convolution layer, the resulting node embeddings were passed through a linear classifier to obtain class scores for each node.

\begin{equation}
\text{Z} = \text{H}^{(L)}\text{W}_{\mathrm{out}} + \text{b}_{\mathrm{out}} \label{Eq:MLP}
\end{equation}

where \(\text{H}^{(L)}\) is the final node embedding matrix, \(\text{W}_{\mathrm{out}}\) and \(\text{b}_{\mathrm{out}}\) are the classifier parameters, and \(\text{Z}\) contains the logits for the three limitation classes. The predicted class label for node \(j\) was then obtained as

\begin{equation}
\hat{y}_j = \arg\max_{c \in \{0,1,2\}} Z_{j,c} \label{Eq:Softmax}
\end{equation}

\subsubsection{Graph attention network framework (GAT)}

To incorporate adaptive neighborhood weighting, we used a graph attention network (GAT) for node classification on the CO$_2$ response curve \citep{Petar2018} (\figpanels{Fig:Graph_Model_Frameworks}{b}). Each curve was represented as a graph $G = (\mathcal{V}, \mathcal{E})$, where node features were defined as in Eq.~\ref{Eq:NodeLabel}. The graph connectivity was constructed using the bidirectional kNN graph and the ASG connectivity scheme (Section~\ref{Sec:EdgeAttr}). In addition, the GAT framework allows for the explicit incorporation of edge attributes, computed for each edge as described in Eq.~\ref{Eq:Edgelabel}. Thus, edge attributes provide additional information on pairwise changes in auxiliary signals.

The GAT model consisted of stacked graph attention layers followed by batch normalization, \(tanh\) activation, and dropout (\(p=0.2\)). For the intermediate graph attention layers, the updated node representation is computed by concatenating the outputs from all \((K = 5)\) attention heads:
\begin{equation}
h_j^{(\ell+1)} =
\mathop{\Vert}_{k=1}^{K}
\sigma\left(
\sum_{q \in \mathcal{N}(j)}
\alpha_{jq}^{(\ell,k)} \mathrm{W}^{(\ell,k)} \mathrm{h}_q^{(\ell)}
\right)
\end{equation}

where \(\sigma(\cdot)\) denotes the nonlinear transformation. For the final graph attention layer, the head-wise outputs are averaged:
\begin{equation}
h_j^{(L)} =
\frac{1}{K}
\sum_{k=1}^{K}
\sigma\left(
\sum_{q \in \mathcal{N}(j)}
\alpha_{jq}^{(L-1,k)} \mathrm{W}^{(L-1,k)} \mathrm{h}_q^{(L-1)}
\right)
\end{equation}

Finally, the node embeddings from the last GAT layer are passed through a linear classifier as described for the GCN baseline \Eqref{Eq:MLP}, \Eqref{Eq:Softmax} to obtain node-wise class scores and the predicted limitation label for each node. 

\subsubsection{Graph U-Net framework}

To capture both local and coarse-scale structure along the CO$_2$ response curve, we used a Graph U-Net-based node classifier (\figpanels{Fig:Graph_Model_Frameworks}{c}). Each curve was represented as a graph \(G = (\mathcal{V}, \mathcal{E})\) \citep{Gao2019}, with connectivity defined using either the kNN scheme or the ASG scheme described in Section~\ref{Sec:EdgeAttr}. In the GCN-U-Net formulation, message passing was based only on the graph connectivity and node features, whereas in the GAT-U-Net formulation, the hierarchical attention layers additionally incorporated edge attributes during message passing.

The model consisted of an encoder–decoder Graph U-Net architecture of depth = 2, followed by batch normalization, \(tanh\) activation, dropout (\(p=0.2\)), and a final linear classifier. Each encoder level consisted of a graph convolution or graph attention layer, followed by top-$k$ pooling with a pooling ratio of 0.8, which sequentially eliminated nodes and edges, enabling the model to learn coarser structural features. In the decoder path, the graphs were restored to the original structure using the saved pooling indices, and the encoder features were merged using residual skip connections. This hierarchical update can be written schematically as
\begin{equation}
\text{H}^{(\ell+1)} =
\mathrm{Unpool}\!\left(
\mathrm{Pool}\!\left(
\mathrm{Conv}\!\left(\text{H}^{(\ell)}, \hat{\text{A}}\right)
\right)
\right)
\end{equation}
In practice, the GCN-U-Net used the PyTorch Geometric \textit{GraphUNet} module; therefore, edge attributes were not used directly in this variant. The GAT-U-Net used a custom encoder--decoder implementation with a graph attention network and \textit{TopKPooling} \citep{Lenssen2019}. During pooling, both the edge index and edge attributes were filtered to the retained subgraph. During decoding, the saved encoder-level edge index and edge attributes were reused after unpooling. Edge attributes were not recomputed after pooling or unpooling. Finally, the node embeddings were mapped to class scores as described for the GCN baseline and GAT \Eqsref{Eq:MLP}{Eq:Softmax}, yielding node-wise class scores and the predicted limitation label for each node. Training Graph U-Net variants for a larger number of epochs occasionally led to unstable optimization dynamics, including episodes of gradient explosion. Consequently, unlike the other models, the Graph U-Net variants were restricted to 600 training epochs.

\subsubsection{Model training}

All models were trained for the same node-level classification task. To ensure a fair comparison across model families, the synthetic dataset was partitioned at the curve level into training (60\%), validation (20\%), and test subsets (20\%). The training subset was used for learning and model weights and biases, the validation subset for hyperparameter tuning, and the held-out test subset for final evaluation and model comparison.

Prior to model fitting, node features and edge attributes were normalized using z-score normalization with mean and standard deviation calculated on the training set and applied consistently across the training, validation, and test subsets. For the deep learning baseline, each curve was represented as an ordered sequence of pointwise features. For graph-based models, each curve was represented as a graph with node features $\mathbf{x}_j$ \Eqref{Eq:NodeLabel}, and edge features $\mathbf{e}_{uv}$ \Eqref{Eq:Edgelabel}.

All models were optimized using the Adam optimizer. Model performance (macro F1-score) was monitored on both the training and validation subsets across 800–1000 epochs, and the best validation performance was retained for final testing \citep{Kar2024}. To ensure a statistically sound comparison across models, 30 independent model training instances were run, each with unique random initializations. The learning rate and weight decay were set to 0.001 and 0.0001, respectively, with a dropout rate of 0.2 to avoid overfitting \citep{Srivastava2014,Kar2024}. A batch size of 128 was used throughout training; this choice was appropriate given the smaller subset of A--Ci curves (1250 curves) for each curve-length-specific deep learning model. The remaining model-specific hyperparameters were selected through hyperparameter tuning on the validation set (Table~S2). Unless stated, the selected design was used for all model training, attributing performance differences primarily to the model structure rather than to different hyperparameter choices.

Three loss functions were compared: standard cross-entropy ($\mathcal{L}_{\mathrm{CE}}$), weighted cross-entropy ($\mathcal{L}_{\mathrm{WCE}}$), and weighted focal loss ($\mathcal{L}_{\mathrm{WF}}$) \citep{Elkan2001}, \citep{He2009}, \citep{Lin2017}. The standard multiclass cross-entropy loss was written as
\begin{equation}
\mathcal{L}_{\mathrm{CE}} =
- \sum_{j=1}^{N_T}\sum_{c=1}^{C}
y_{jc}\log(\hat{y}_{jc}),
\end{equation}
where $N_T$ is the number of nodes in the training set, $y_{jc}$ is the binary indicator for the $j^{th}$ node's classification, and $\hat{y}_{jc}$ is the predicted probability for that class. To account for class imbalance among the limiting states, a weighted class cross-entropy loss was also considered:
\begin{equation}
\mathcal{L}_{\mathrm{WCE}} =
- \sum_{j=1}^{N_T}\sum_{c=1}^{C}
w_c\, y_{jc}\log(\hat{y}_{jc}),
\end{equation}
where $w_c$ denotes the class-specific weight estimated from the class distribution in the training set. In addition, focal loss was used in selected experiments to place higher emphasis on hard-to-classify nodes:
\begin{equation}
\mathcal{L}_{\mathrm{WF}} =
- \sum_{j=1}^{N_T}\sum_{c=1}^{C}
w_c (1-\hat{y}_{jc})^{\gamma}
y_{jc}\log(\hat{y}_{jc}),
\end{equation}
where $\gamma$ is the focusing parameter. When $\gamma = 0$, \(\mathcal{L}_{\mathrm{WF}}\) reduces to \(\mathcal{L}_{\mathrm{WCE}}\).

\subsubsection{Model evaluation}

Comparison of the proposed model frameworks was performed on the 20\% held-out test set. This focused on the effect of architectural choices, graph connectivity, attention, hierarchical pooling, and weighted loss functions on node-wise limitation-state classification. Table~\ref{Tab:scenario_summary} shows the model scenarios evaluated in this study. Quantitative evaluation was performed using classification-based metrics: accuracy, precision, recall, and F1-score \citep{Kar2024}.

Since the task involved three limitation classes, precision, recall, and F1-score were computed using a one-vs-rest formulation for each class and then macro-averaged across classes. Let \(C\) denote the number of classes and \(N_T\) denote the total number of test nodes. For class \(c \in \{1,\ldots,C\}\), let \(TP_c\), \(FP_c\), and \(FN_c\) denote the true positives, false positives, and false negatives for that class, respectively. The class-wise precision, recall, and F1-score were computed as
\begin{equation}
\mathrm{Precision}_c =
\frac{TP_c}{TP_c + FP_c},
\end{equation}
\begin{equation}
\mathrm{Recall}_c =
\frac{TP_c}{TP_c + FN_c},
\end{equation}
and
\begin{equation}
\mathrm{F1}_c =
\frac{2 \cdot \mathrm{Precision}_c \cdot \mathrm{Recall}_c}
{\mathrm{Precision}_c + \mathrm{Recall}_c}.
\end{equation}

The reported macro-averaged metrics were then calculated as
\begin{equation}
\mathrm{Precision}_{\mathrm{macro}} =
\frac{1}{C}\sum_{c=1}^{C} \mathrm{Precision}_c,
\end{equation}
\begin{equation}
\mathrm{Recall}_{\mathrm{macro}} =
\frac{1}{C}\sum_{c=1}^{C} \mathrm{Recall}_c,
\end{equation}
and
\begin{equation}
\mathrm{F1}_{\mathrm{macro}} =
\frac{1}{C}\sum_{c=1}^{C} \mathrm{F1}_c.
\end{equation}

Accuracy was computed as the fraction of correctly classified test nodes: \begin{equation} \mathrm{Accuracy} = \frac{1}{N_T}\sum_{j=1}^{N_T} \mathbb{I}\left(\hat{y}_j = y_j\right), \end{equation} where \(y_j\) and \(\hat{y}_j\) denote the true and predicted labels of test node \(j\), respectively, and \(\mathbb{I}(\cdot)\) is the indicator function.

The main metric is the macro-F1-score, since the three limitation classes were not evenly distributed, and accuracy alone can exaggerate performance when the model prioritizes the dominant class over the minority or the transition points between classes \citep{George2010}. Accuracy, precision, and recall were presented to provide further insight into overall correctness and class-specific behavior.

The evaluation was organized in two stages. First, the feature-based baseline, namely RF, SVM, XGBoost, and the DL baseline, were compared against the baseline graph neural network model (GCN), in order to assess the value of introducing graph connectivity for node-wise limitation-state classification. Second, after establishing this baseline graph advantage, more detailed comparisons were carried out within the graph-based frameworks by examining the effects of graph density, attention-based aggregation, hierarchical Graph U-Net pooling, and the imbalance-aware loss functions $\mathcal{L}_{\mathrm{CE}}$, $\mathcal{L}_{\mathrm{WCE}}$, and $\mathcal{L}_{\mathrm{WF}}$.

\subsubsection{Benchmark label extraction, post-training analysis, and model interpretation}

A recently published automated fitting-based model, PhoTorch, was used to extract the limiting states. This was done as a post-processing step, since PhoTorch performs parameter optimization and limiting-state identification simultaneously \citep{Lei2025}. PhoTorch was fitted to the same noisy, subsampled A--Ci curves used for model evaluation, using the observed $C_\mathrm{i}$, $A_{\mathrm{net}}$, and the same environmental assumptions used in data generation, including $T_{\mathrm{leaf}}$ and O$_\mathrm{2}$. Each curve was fitted independently, with provision to fit $R_\mathrm{d}$ and $g_\mathrm{m}$ along with the photosynthetic parameters, without light-response fitting, and with fixed values for $\Gamma^\ast$, $K_\text{o}$, $K_\text{c}$, and the $R_\mathrm{d}$ ratio. Fitted $A_c$, $A_j$, and $A_p$ rates at each observed point were used to extract the pointwise limiting state. The limitation state was set based on the minimum of these three rates. These PhoTorch-derived labels were then evaluated using the same evaluation metrics as the other machine learning models.

To further analyze model behavior after training, two complementary post-training analyses were performed. First, \textit{GNNExplainer} was used to qualitatively examine the node-wise predictions of the graph-based models \citep{Ying2019}. For a selected target node, the explainer identifies the most important connections and the corresponding subgraph that contributes to the model prediction. Following this, an edge-level ablation study was performed by perturbing the identified important edges and measuring the resulting change in class-probability contrast, thereby highlighting the connections that lead to the correct classification of the limitation state. The explanations were generated after model training only for visualization and interpretation; they were not used during model optimization or model selection. 

Second, a statistical analysis of the 12 graph-attention-based model configurations was performed using 30 independent model training runs per configuration. The analysis includes a one-way analysis of variance (ANOVA) \citep{Fisher1992} to test for statistically significant differences among models. When ANOVA showed a significant difference, Tukey’s honestly significant difference (Tukey–HSD) was used to identify pairwise differences among model configurations \citep{Tukey1949,Kar2020}. The ANOVA and Tukey–HSD results were used to assign models to statistically similar or distinct performance groups, reflecting the stability of the training runs across repeated model initializations. Altogether, these analyses provided insight into the mechanism of the graph predictions and statistical evidence of variations in overall model performance.

\section{Results and discussion}

\subsection{The limitations of feature-based curve-level classification: Establishing the baseline}

\subsubsection{Performance of the ML and NN baselines}

\begin{figure}
    \centering
    \includegraphics[width=0.5\textwidth]{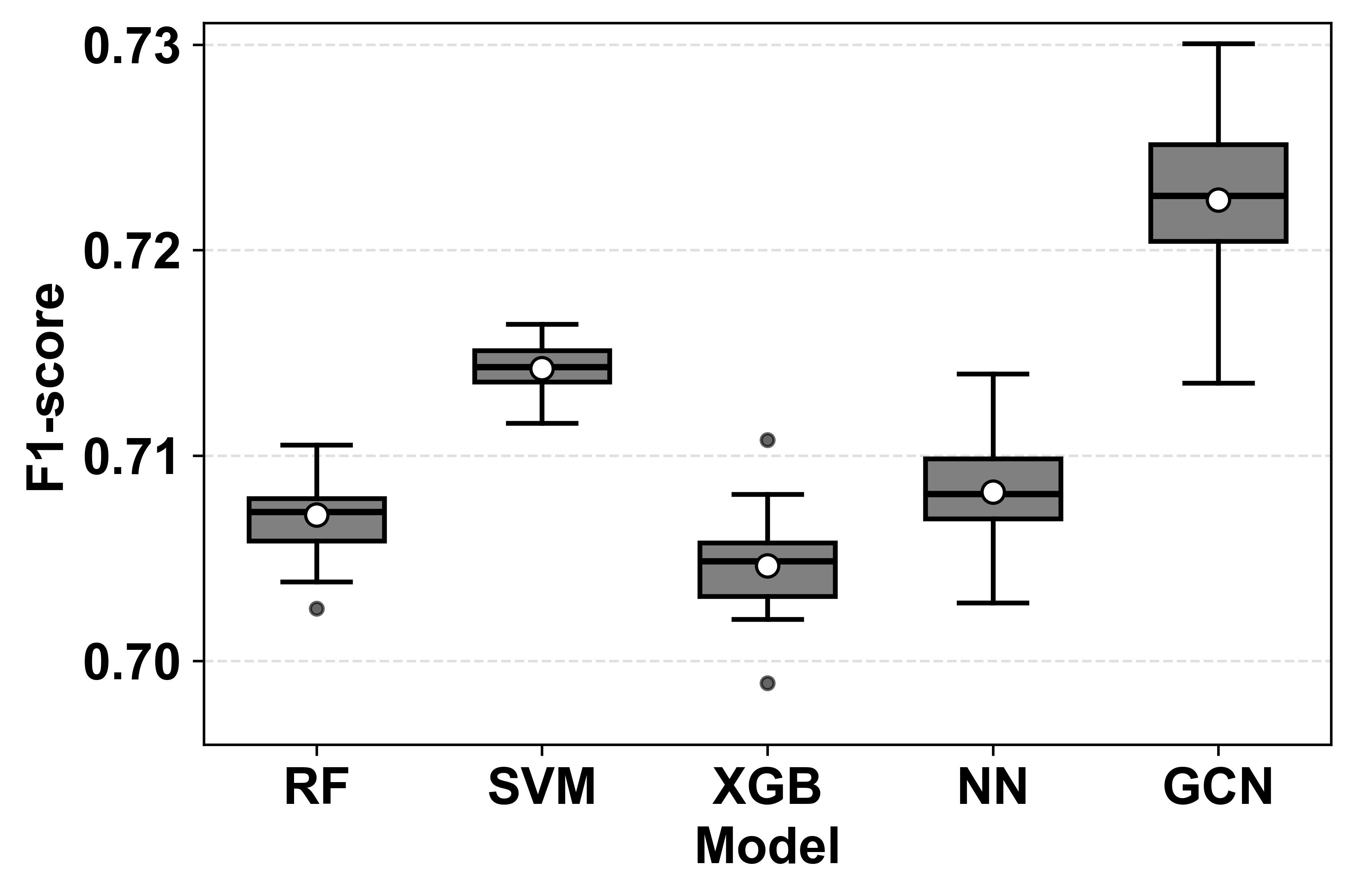}
    \caption{Feature-based and GCN baseline performance comparison. F1-score distributions across RF, SVM, XGB, NN, and GCN models over 30 runs. Box plots show median, spread, and a white circle indicating the mean values.}
    \label{Fig:ML_models_comp}
\end{figure}

The conventional machine learning baselines showed broadly similar performance, with SVM as the best among the three feature-based models, with an F1-score of \(0.714 \pm 0.001\) (Fig.~\ref{Fig:ML_models_comp}). Random Forest and XGBoost yielded F1-scores of \(0.707 \pm 0.002\) and \(0.705 \pm 0.002\), respectively (Fig.~\ref{Fig:ML_models_comp}). These results indicate that standard feature-based classifiers cannot recover a significant part of pointwise limitation-state information from node features \Eqref{Eq:NodeLabel} alone. 

This limitation can also be observed in the representative A--Ci curve shown in Fig.~S2(a) and Table~S3. While the feature-based models reproduce the broad sequence from $A_c$- to $A_j$- and $A_p$-limited regions, they struggle to replicate the correct transitions between limiting processes (Fig.~S2(b)). In all cases, the shift from $A_c$ to $A_j$-limited region is delayed, while the transition into the $A_p$-limited region is too early (Fig.~S2(b)). The ensemble feed-forward neural network produced a performance level similar to that of the conventional machine learning baselines, with an F1-score of $0.708 \pm 0.002$ (Fig.~\ref{Fig:ML_models_comp}). Although the NN model can learn nonlinear interactions across the ordered feature set of a curve, it is limited since the input is treated only as a flattened sequence of pointwise measurements.

\subsubsection{Limitations of feature-based models}

The common limitation of the pointwise ML and NN baselines is not the complete absence of curve-level information, since the full A--Ci curve is provided to the models during prediction. Rather, these models represent the curve as a fixed tabular or flattened input, without explicitly encoding the relational structure among measurement points. Consequently, although they can learn global patterns across the curve, they are not directly informed about how neighboring measurements are connected along the underlying photosynthetic response trajectory. They also do not propagate information through predefined physiological or geometric relationships among points.

This limitation is reflected in the predicted label sequences shown in the representative curve in Fig.~S2(a). The feature-based models contained missclassified insertions of one limiting class into another (Fig.~S2(b)). The NN model misclassified two points in the $A_j$-limited region as $A_c$- and $A_p$-limited, and XGBoost inserted an $A_p$-limited point within the $A_j$-limited region (Fig.~S2(b)). Such insertions indicate that individual points can be misclassified in isolation when the relational context is not represented. Overall, these results show that predictive flexibility alone is insufficient unless the relationship between neighboring points is explicitly encoded.

\begin{figure*}
    \centering
    \includegraphics[width=\textwidth]{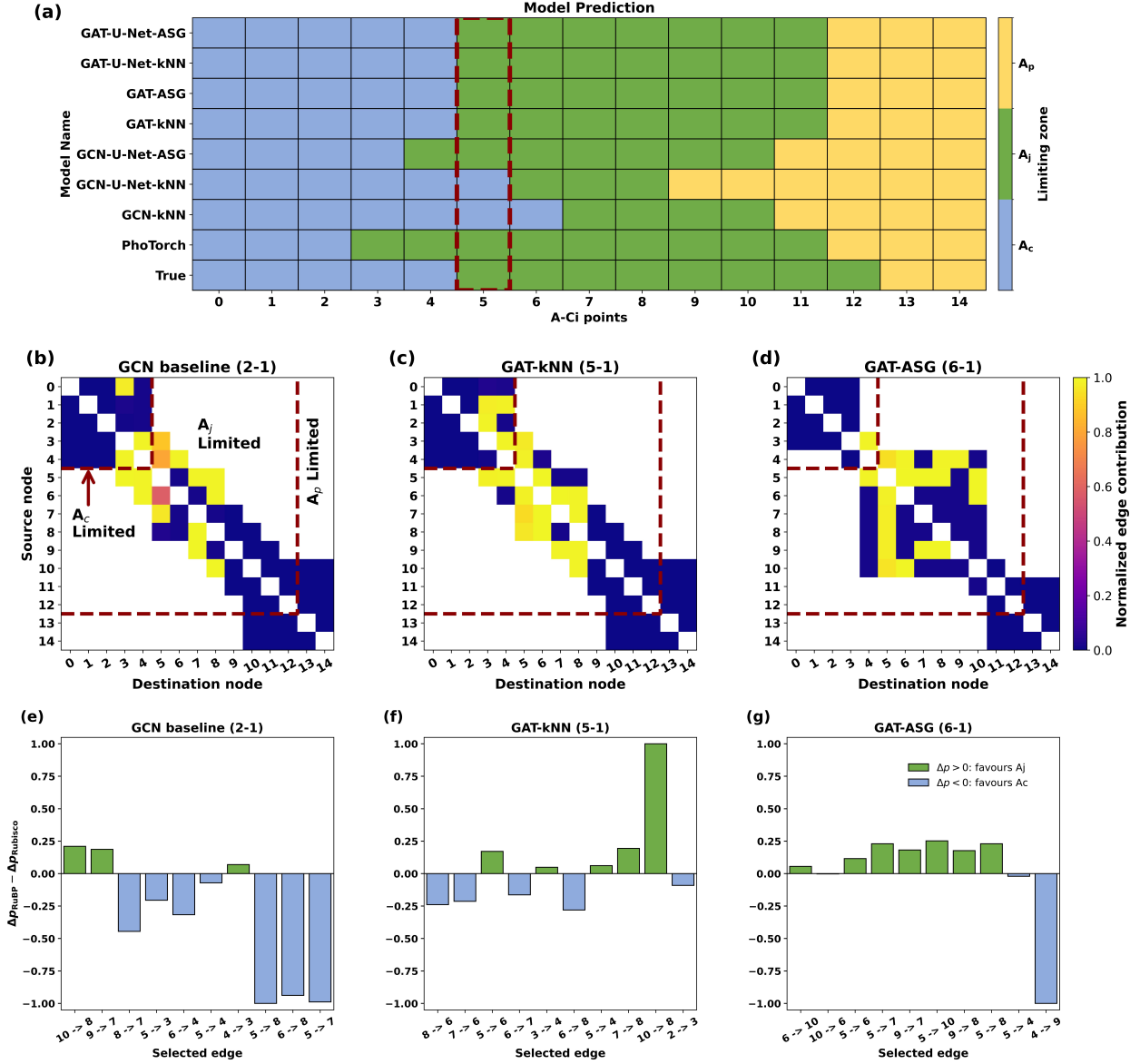}
    \caption{Prediction, explanation, and ablation results for a representative A--Ci curve. 
    (a) True and predicted limiting zones for the selected model configurations, with the dashed rectangle marking the transition region around node 5. (b)--(d) GNNExplainer-derived edge-importance matrices for node 5 for the GCN baseline, GAT-kNN, and GAT-ASG models. (e)--(g) Edge-ablation effects for the top 10 selected important edges, shown as the normalized change in class-probability contrast, \(\Delta p_{\mathrm{Aj}}-\Delta p_{\mathrm{Ac}}\).} \label{Fig:GNN_Edge_Ablation}
\end{figure*}

\begin{table*}[!t]
\centering
\caption{Summary of model categories and evaluation metrics.}
\label{Tab:model_summary}
\vspace{0.1cm}
\begin{tabular}{l p{3.2cm} c l c c c c}
\hline
\multirow{2}{*}{\textbf{ID}} & \multirow{2}{=}{\textbf{Model category}} 
& \multicolumn{2}{c}{\textbf{Model}} & \multirow{2}{*}{\textbf{F1-score}} & \multirow{2}{*}{\textbf{Recall}} & \multirow{2}{*}{\textbf{Precision}} & \multirow{2}{*}{\textbf{Accuracy}} \\
\cline{3-4}
& & \textbf{Code} & \textbf{Name} & & & & \\ \hline
0 & Numerical model & - & \textit{PhoTorch} & 0.735 & 0.753 & 0.751 & 0.751 \\ \hline
1 & DL baseline & 1-1 & NN & 0.713 & 0.715 & 0.714 & 0.766 \\ \hline
2 & GNN baseline & 2-1 & GCN-kNN-$\mathcal{L}_{\mathrm{CE}}$ & 0.722 & 0.726 & 0.723 & 0.778 \\ \hline
\multirow{3}{*}{3} 
& \multirow{3}{=}{Graph U-Net + kNN} 
& 3-1 & GCN-U-Net-kNN-$\mathcal{L}_{\mathrm{CE}}$     
& 0.780 & 0.783 & 0.777 & 0.819 \\
&  & 3-2 & GCN-U-Net-kNN-$\mathcal{L}_{\mathrm{WCE}}$  
& 0.764 & 0.769 & 0.762 & 0.800 \\
&  & 3-3 & GCN-U-Net-kNN-$\mathcal{L}_{\mathrm{WF}}$ 
& 0.768 & 0.774 & 0.766 & 0.803 \\
\hline
\multirow{3}{*}{4} 
& \multirow{3}{=}{Graph U-Net + ASG} 
& 4-1 & GCN-U-Net-ASG-$\mathcal{L}_{\mathrm{CE}}$     
& 0.817 & 0.819 & 0.815 & 0.851 \\
&  & 4-2 & GCN-U-Net-ASG-$\mathcal{L}_{\mathrm{WCE}}$ 
& 0.814 & 0.818 & 0.812 & 0.844 \\
&  & 4-3 & GCN-U-Net-ASG-$\mathcal{L}_{\mathrm{WF}}$ 
& 0.814 & 0.817 & 0.811 & 0.846 \\ 
\hline
\multirow{3}{*}{5} 
& \multirow{3}{=}{\textbf{Graph + Attention + kNN}} 
& 5-1 & GAT-kNN-$\mathcal{L}_{\mathrm{CE}}$     
& 0.854 & 0.855 & 0.854 & 0.883 \\
&  & \textbf{5-2} & \textbf{GAT-kNN-}$\mathbf{\mathcal{L}}_{\mathrm{\textbf{WCE}}}$  
& \textbf{0.857} & \textbf{0.861} & \textbf{0.853} & \textbf{0.882} \\
&  & 5-3 & GAT-kNN-$\mathcal{L}_{\mathrm{WF}}$ 
& 0.855 & 0.860 & 0.851 & 0.881 \\
\hline
\multirow{3}{*}{6} 
& \multirow{3}{=}{Graph + Attention + ASG} 
& 6-1 & GAT-ASG-$\mathcal{L}_{\mathrm{CE}}$     
& 0.847 & 0.848 & 0.846 & 0.877 \\
&  & 6-2 & GAT-ASG-$\mathcal{L}_{\mathrm{WCE}}$  
& 0.846 & 0.851 & 0.842 & 0.872 \\
&  & 6-3 & GAT-ASG-$\mathcal{L}_{\mathrm{WF}}$ 
& 0.844 & 0.850 & 0.841 & 0.871 \\
\hline
\multirow{3}{*}{7} 
& \multirow{3}{=}{Graph U-Net + Attention + kNN} 
& 7-1 & GAT-U-Net-kNN-$\mathcal{L}_{\mathrm{CE}}$     
& 0.850 & 0.853 & 0.849 & 0.879 \\
&  & 7-2 & GAT-U-Net-kNN-$\mathcal{L}_{\mathrm{WCE}}$  
& 0.852 & 0.857 & 0.849 & 0.878 \\
&  & 7-3 & GAT-U-Net-kNN-$\mathcal{L}_{\mathrm{WF}}$ 
& 0.845 & 0.849 & 0.842 & 0.873 \\
\hline
\multirow{3}{*}{8} 
& \multirow{3}{=}{Graph U-Net + Attention + ASG} 
& 8-1 & GAT-U-Net-ASG-$\mathcal{L}_{\mathrm{CE}}$     
& 0.851 & 0.853 & 0.850 & 0.881 \\
&  & 8-2 & GAT-U-Net-ASG-$\mathcal{L}_{\mathrm{WCE}}$ 
& 0.849 & 0.853 & 0.847 & 0.877 \\
&  & 8-3 & GAT-U-Net-ASG-$\mathcal{L}_{\mathrm{WF}}$ 
& 0.850 & 0.855 & 0.847 & 0.876 \\
\hline
\end{tabular}
\end{table*}

\subsection{The significance of relational context: Introducing graph connectivity}
\subsubsection{Performance of the GCN framework}

Introducing graph connectivity improves node-wise classification beyond pointwise baselines, indicating that relational context provides useful information for the limitation-state classification problem. The baseline GCN model with kNN connectivity and cross-entropy loss achieved an F1-score of \(0.722 \pm 0.003\), outperforming all ML and DL feature-based models (Fig.~\ref{Fig:ML_models_comp}). This result shows that even a simple graph convolution framework can improve classification by representing the A--Ci curve as a connected structure rather than as a collection of independent points. Compared with feature-based models, the GCN-kNN model better preserves the overall progression of the limiting states along the A--Ci trajectory. In particular, the GCN-kNN model shows the fewest misclassifications among the \(A_j\) limitations compared to the ML and DL models (\figpanels{Fig:GNN_Edge_Ablation}{a}). 

However, the GCN-kNN result also shows that graph connectivity alone does not completely resolve the limitation-state assignment problem. Although GCN-kNN improved on the feature-based baselines for ML and DL, its F1-score of \(0.722\) remained lower than that of the automated fitting-based benchmark PhoTorch, which achieved an F1-score of \(0.735\), although GCN-kNN achieved a higher overall accuracy of \(0.778\) compared to \(0.751\) for PhoTorch (Table~\ref{Tab:model_summary}). Qualitative predictions show that both methods still make errors in the transition region. PhoTorch shifts the onset of the \(A_j\)-limited region earlier, while GCN-kNN delays the \(A_c \rightarrow A_j\) transition and also predicts the onset of the \(A_p\)-limited region too early \figpanels{Fig:GNN_Edge_Ablation}{a}. Thus, while the GCN demonstrates the value of representing the A--Ci curve as a connected graph, its performance relative to the automated-fitting-based benchmark indicates that fixed-neighborhood aggregation remains insufficient to resolve biochemical transition ambiguity completely.

\subsubsection{Contributions of graph connectivity}

The advantage of graph connectivity is not only that it improves aggregate performance, but also that it changes how curve-level information is used for node classification. In the ML and NN baselines, the full A--Ci curve is provided to the model, allowing each prediction to depend on the entire curve. However, these representations remain tabular or flattened and do not contain the information about the connections between measurement points. In contrast, the GCN uses information from neighboring nodes to predict a target node’s class using message passing between connected neighbors. This is particularly important for limiting region identification in the A--Ci curves, since the adjacent points often belong to the same biochemical limiting region or lie close to a transition boundary.

This is evident from the edge-importance map for the classification of node 5 in the representative curve (Fig.~S2(a) \& \figpanels{Fig:GNN_Edge_Ablation}{b}). The dominant edges contributing to the classification of the selected transition node are concentrated among nearby nodes around the \(A_c \rightarrow A_j\) boundary, rather than arbitrarily distributed across the curve (\figpanels{Fig:GNN_Edge_Ablation}{b}). Thus, the GCN does not rely only on a flattened curve representation but interprets the target point through its explicitly connected local response trajectory. In this context, graph connectivity plays a physiologically meaningful role, enabling neighboring nodes to reinforce signals for limiting-state identification and reduce the instances of isolated missclassification insertions. At the same time, the mismatch in the transition regions indicates that simple neighborhood aggregation is useful, but not sufficient.

\begin{figure*}
    \centering
    \includegraphics[width=\textwidth]{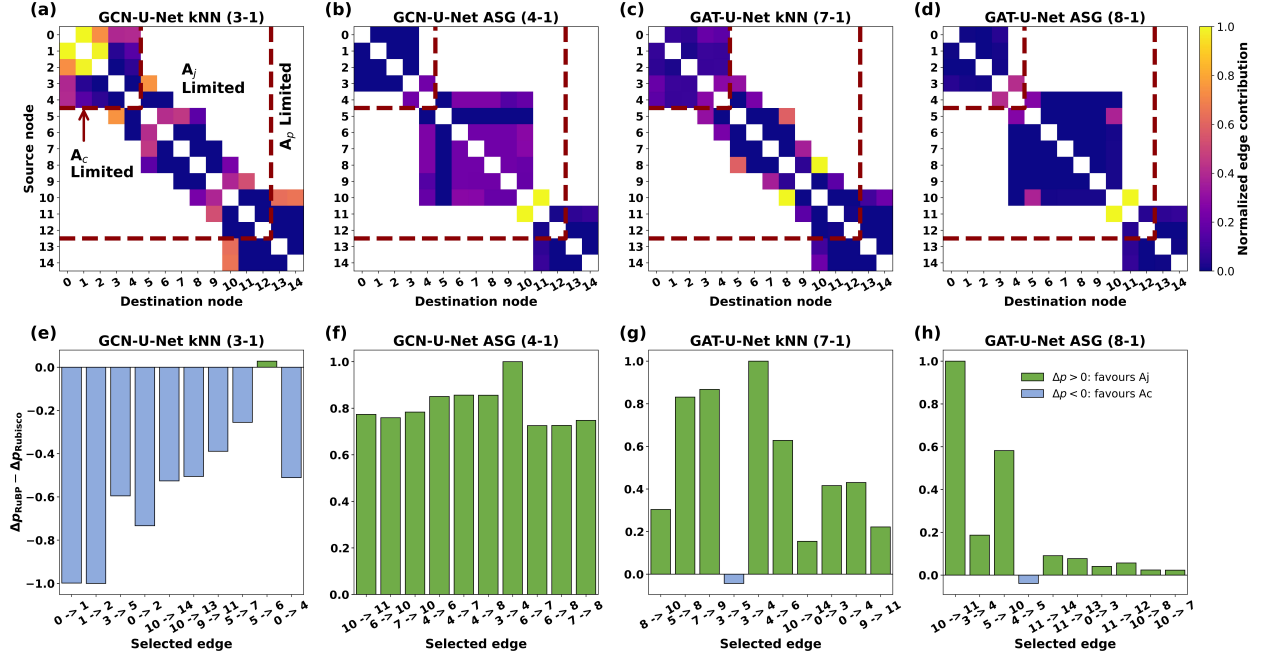}
	\caption{Graph U-Net edge importance and ablation analysis on the representative A--Ci curve. (a)--(d) GNNExplainer-derived edge-importance matrices for node 5 for GCN-U-Net-kNN, GCN-U-Net-ASG, GAT-U-Net-kNN, and GAT-U-Net-ASG, respectively. (e)--(h) Edge-ablation effects for the top 10 selected important edges, shown as the normalized change in class-probability contrast, \(\Delta p_{\mathrm{Aj}}-\Delta p_{\mathrm{Ac}}\).} \label{Fig:GNN_UNet_Edge_Ablation} 
\end{figure*}

\subsection{Resolving ambiguity at regime transitions: Power of attention}

\subsubsection{Attention-based aggregation improves over fixed graph convolution}

All GAT variants outperformed both the baseline GCN-kNN model and the automated fitting-based PhoTorch benchmark on all evaluation metrics (Table~\ref{Tab:model_summary}). This indicates that attention-based neighborhood aggregation provides a stronger representation of the A--Ci curve structure for biochemical limitation-state identification. \figpanels{Fig:GNN_Edge_Ablation}{a} highlights the qualitative improvement in GAT predictions, which better recover the true limiting state than the GCN baseline and PhoTorch. In particular, the GAT variants resolve the \(A_c \rightarrow A_j\) transition, while also maintaining a more stable continuation through the \(A_j\)-limited region (\figpanels{Fig:GNN_Edge_Ablation}{a}).

In the kNN setting, the best performance was obtained with GAT-kNN-\(\mathcal{L}_{\mathrm{WCE}}\), which achieved the highest F1-score of \(0.857\) among the standard GAT models (Table~\ref{Tab:model_summary}). The corresponding cross-entropy and weighted focal-loss variants achieved similar F1-scores of \(0.854\) and \(0.855\), respectively (Table~\ref{Tab:model_summary}). In the ASG setting, the best model was GAT-ASG-\(\mathcal{L}_{\mathrm{CE}}\), with an F1-score of \(0.847\), while the weighted cross-entropy and weighted focal-loss variants achieved F1-scores of \(0.846\) and \(0.844\), respectively (Table~\ref{Tab:model_summary}). Thus, the effect of the loss formulation was relatively small within each connectivity setting, although weighted cross-entropy slightly improved recall in both the kNN and ASG graph configuration (Table~\ref{Tab:model_summary}). 

The GNNExplainer and edge-ablation analyses further clarify why the GAT variants improve over the GCN baseline. For the selected transition point (A--Ci point = 5, \figpanels{Fig:GNN_Edge_Ablation}{a}), the edge-importance matrix reveals that the GCN baseline assigns low importance to incoming edges from neighboring measurements toward the transition point, while this is not the case in the GAT variants (\figpanels{Fig:GNN_Edge_Ablation}{b--d}). Additionally, the ablation study of the top 10 important edges reveals that these edges predominantly support the incorrect \(A_c\)-limited classification (\figpanels{Fig:GNN_Edge_Ablation}{e}). This ambiguity is also reflected in the predicted class probabilities: GCN assigns a substantial probability to the true \(A_j\)-limited class (\(p(A_j) = 0.341\)), but still gives the highest probability to the incorrect \(A_c\)-limited class (\(p(A_c) = 0.634\)) (Table~S4).

In contrast, both GAT-kNN and GAT-ASG assign greater importance to edges that carry information from neighboring measurements into the selected transition region (\figpanels{Fig:GNN_Edge_Ablation}{c,d}). This suggests that attention-based aggregation enables the model to leverage local relational context to resolve ambiguity at the \(A_c \rightarrow A_j\) transition. This additional relational context is reflected in the class probabilities, where GAT-kNN and GAT-ASG assign the highest probability to the correct \(A_j\)-limited class (Table~S4). Edge-ablation analysis provides further support for this interpretation. When the ten most important edges were perturbed, several important GCN edges produced negative values of \(\Delta p_{\mathrm{Aj}}-\Delta p_{\mathrm{Ac}}\), indicating that they favored the incorrect Rubisco-limited prediction (\figpanels{Fig:GNN_Edge_Ablation}{e}). By contrast, the important edges in the GAT variants produced predominantly positive values of \(\Delta p_{\mathrm{Aj}}-\Delta p_{\mathrm{Ac}}\), indicating that these edges supported the correct RuBP-limited classification (\figpanels{Fig:GNN_Edge_Ablation}{f,g}). Thus, the improvement of GAT over GCN arises not only from assigning different importance to neighboring edges, but also from using those edges in a direction that supports the correct biochemical limitation state near the transition region.

\subsubsection{kNN connectivity enables stronger cross-regime information flow than ASG}

\begin{figure*}
    \centering
    \includegraphics[width=\textwidth]{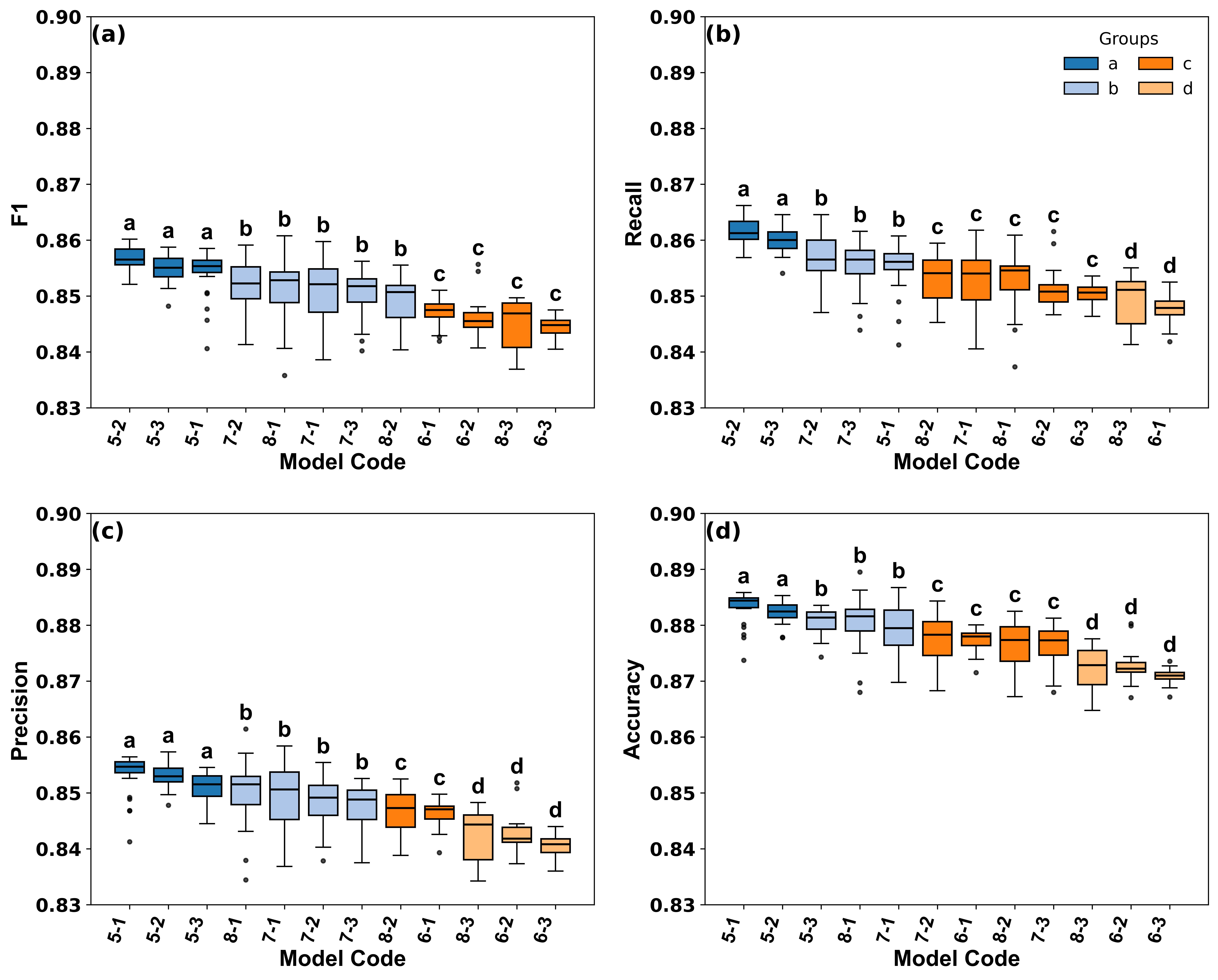}
    \caption{Statistical comparison of GNN model configurations. Comparison of model performance across the 12 GNN configurations using four test metrics: F1-score, recall, precision, and accuracy. Box plots show the distribution of test performance over 30 independent runs, and letters denote statistically distinct Tukey--HSD groups.} \label{Fig:Tukey_HSD}
\end{figure*}

The comparison between the kNN and ASG graph settings shows that graph construction affects both predictive performance and how the model uses neighborhood information (Fig.~\ref{Fig:GNN_Edge_Ablation}). Quantitatively, the standard GAT-kNN variants performed slightly better than the corresponding GAT-ASG variants in all evaluation metrics (Table~\ref{Tab:model_summary}). The Tukey--HSD results also indicate that this performance gap, although small in magnitude, was statistically significant for the standard GAT variants, with the kNN and ASG models assigned to distinct performance groups (Fig.~\ref{Fig:Tukey_HSD}). These results indicate that the simpler proximity-based kNN graph provided a more effective connectivity structure than the ASG graph for node-wise limitation-state classification.

The GNNExplainer-derived edge-importance patterns help explain this difference. In the kNN graph, the important edges for the selected transition point include connections spanning the neighboring \(A_c\)- and \(A_j\)-limited portions of the curve (\figpanels{Fig:GNN_Edge_Ablation}{c}). This allows the attention mechanism to compare and contrast information from both sides of the \(A_c \rightarrow A_j\) transition before assigning the target node’s limiting state. This enhances model training, as models can now be trained to give preference to the correct side of the transition point. In contrast, the ASG graph emphasizes the connectivity within the predefined group and provides only limited communication between different groups (\figpanels{Fig:GNN_Edge_Ablation}{d}). The ASG structure is only helpful when the predefined groups can accurately capture the regime boundaries. In realistic settings, where gas exchange data is noisy, these predefined groups are often inaccurate, and the classification of transition points is negatively impacted by resistance to inter-group message passing.

Furthermore, the edge-ablation results show that the GAT-kNN model provided stronger support for correct classification than the GAT-ASG model, as indicated by the larger positive values of \(\Delta p_{A_j}-\Delta p_{A_c}\) in GAT-kNN compared to GAT-ASG (\figpanels{Fig:GNN_Edge_Ablation}{f--g}). In the GAT-ASG model, several supportive edges originated from the transition node itself. Although this was sufficient for the representative curve, such reliance may be less robust in general because transition-node features are intrinsically ambiguous (\figpanels{Fig:GNN_Edge_Ablation}{g}). Consequently, the superior overall performance of GAT-kNN indicates that permitting broader cross-regime communication is more beneficial than imposing a strongly clustered connectivity structure.

\subsection{Effect of hierarchical graph pooling with Graph U-Net}

\subsubsection{Hierarchical pooling improves GCN but not the best GAT model}

Introducing hierarchical pooling through Graph U-Net improved the convolution-based GCN framework, but it did not outperform the best attention-based GAT-kNN model. For the GCN-U-Net family, the best kNN-based result was obtained by GCN-U-Net-kNN-\(\mathcal{L}_{\mathrm{CE}}\), with an F1-score of \(0.780\), higher than the baseline GCN-kNN-\(\mathcal{L}_{\mathrm{CE}}\), which achieved an F1-score of \(0.722\) (Table~\ref{Tab:model_summary}). The ASG-based GCN-U-Net variants showed stronger performance than the corresponding kNN variants, with GCN-U-Net-ASG-\(\mathcal{L}_{\mathrm{CE}}\) achieving the best result \((\mathrm{F1}=0.817)\) (Table~\ref{Tab:model_summary}). Thus, hierarchical pooling clearly improved the graph convolution framework, particularly when combined with ASG connectivity.

However, this improvement did not carry over to attention-based models. The best GAT-U-Net-kNN model, GAT-U-Net-kNN-\(\mathcal{L}_{\mathrm{WCE}}\), achieved an F1-score of \(0.852\) (Table~\ref{Tab:model_summary}). The GAT-U-Net-ASG variants performed similarly, with the best F1-score of \(0.851\) obtained by GAT-U-Net-ASG-\(\mathcal{L}_{\mathrm{CE}}\) (Table~\ref{Tab:model_summary}). These values were slightly lower than the those of best standard GAT-kNN model, GAT-kNN-\(\mathcal{L}_{\mathrm{WCE}}\), which achieved the highest overall F1-score of \(0.857\) (Table~\ref{Tab:model_summary}). Therefore, while Graph U-Net improved the performance of weaker GCN-based models, the best overall performance was achieved by the simpler GAT-kNN framework without hierarchical pooling.

The effect of the loss formulation was secondary to the effect of model architecture and graph connectivity. In the GCN-U-Net family, standard cross-entropy consistently produced the highest F1-score under both kNN and ASG connectivity, showing that imbalance-aware losses did not improve the convolution-based hierarchical models (Table~\ref{Tab:model_summary}). In the GAT-U-Net kNN variant, GAT-U-Net-kNN-\(\mathcal{L}_{\mathrm{WCE}}\) performed slightly better than the GAT-U-Net-kNN-\(\mathcal{L}_{\mathrm{CE}}\), while in the GAT-U-Net ASG variant, the GAT-U-Net-ASG-\(\mathcal{L}_{\mathrm{CE}}\) still outperformed the other loss function variants (Table~\ref{Tab:model_summary}).

\subsubsection{Interaction between hierarchical pooling, attention, and graph connectivity}

The contrasting behavior of Graph U-Net across the GCN and GAT families suggests that the benefit of hierarchical pooling depends on the base message-passing mechanism. In the GCN-U-Net framework, pooling and unpooling helped compensate for the limited flexibility of fixed graph convolution. This was especially evident under ASG connectivity. The GCN-U-Net-kNN model improved over the baseline GCN. However, it still misclassified the target transition point in the representative curve (\figpanels{Fig:GNN_Edge_Ablation}{a}). GCN-U-Net-ASG model correctly classified the same point with the highest probability to the true limiting class (Table~S4). The graph explanation results further support this behavior: under the GCN-U-Net-kNN, many of the important edges were from the \(A_c\)-limited region, which produced negative values of \(\Delta p_{A_j}-\Delta p_{A_c}\) in the ablation study, resulting in incorrect classification (\figpanels{Fig:GNN_UNet_Edge_Ablation}{a,e}). Under GCN-U-Net-ASG, the influential edges were mostly from the \(A_j\)-limited region and produced positive values of \(\Delta p_{A_j}-\Delta p_{A_c}\), supporting the correct \(A_j\)-limited classification (\figpanels{Fig:GNN_UNet_Edge_Ablation}{b,f}). Thus, the hierarchical GCN-U-Net model significantly improved the information exchange within the predefined groups in the ASG graph setting.

However, the same advantage was not observed when the hierarchical pooling was combined with attention-based GNNs. In both GAT-U-Net-kNN and GAT-U-Net-ASG, the selected influential edges mostly supported the correct \(A_j\)-limited classification of the transition point, as indicated by positive values of \(\Delta p_{A_j}-\Delta p_{A_c}\) \figpanels{Fig:GNN_UNet_Edge_Ablation}{c--d,g--h}. This suggests that attention alone is sufficient for the GNN to identify useful neighboring information to resolve transition points. Once this adaptive edge weighting was available, hierarchical pooling did not add a clear advantage. Instead, pooling might have diminished the value of preserving fine-scale local differences across limitation regimes, which are important for node-wise classification, especially at transition points.

This explains the overall best performance of GAT-kNN rather than the GAT-U-Net variants (Table~\ref{Tab:model_summary}). Since kNN connectivity provides direct local connections between neighboring points, the attention mechanism can assign higher weights to the most informative edges within the correct class. This combination allows the model to compare neighbor candidate edges for classifying the ambiguous transition point without first compressing the graph into a pooled representation. In contrast, Graph U-Net introduces an additional hierarchical pooling step, which can be useful for weaker convolution-based models, but is not necessary when attention already provides the flexibility for local aggregation. 

\subsection{Comparative discussion across the attention-based model frameworks}

\subsubsection{Overall comparison of GAT and GAT-U-Net variants}

The Tukey--HSD comparison across the attention-based models shows that the GAT-kNN family achieved the strongest overall performance (Fig.~\ref{Fig:Tukey_HSD} and Table~S5). Among all configurations, GAT-kNN-\(\mathcal{L}_{\mathrm{WCE}}\) achieved the best balance across the four evaluation metrics. This model, hereafter referred to as \textit{SEAGAN} (\textit{domain-Specific and Edge-Aware Graph Attention Network for Dynamic Plant Processes}), has the highest F1-score \((0.857)\) and recall \((0.861)\), while also maintaining high precision \((0.853)\) and accuracy \((0.882)\) (Table~\ref{Tab:model_summary}). Although GAT-kNN-\(\mathcal{L}_{\mathrm{CE}}\) achieved the highest mean precision and accuracy, the larger number of low-performing outliers indicates possible sensitivity to initialization or run-to-run variability (\figpanels{Fig:Tukey_HSD}{c--d}). In contrast, SEAGAN showed a more balanced distribution across the metrics, with top-group membership in all metrics and the highest F1-score and recall (\figpanels{Fig:Tukey_HSD}{a--b}).

The Tukey--HSD analysis shows that GAT-kNN variants consistently rank first across the evaluation metrics. In contrast, the GAT-ASG variants are mostly among the weaker statistical groups (Fig.~\ref{Fig:Tukey_HSD} and Table~S5). This further demonstrates that kNN connectivity provided a more effective graph structure than ASG for the attention-based models, and that its advantage was not limited to a single metric but was reflected in both class recovery and overall prediction accuracy. The comparison also shows that hierarchical pooling did not provide an additional advantage once attention-based aggregation was introduced (Table~\ref{Tab:model_summary}). The GAT-U-Net variants, while competitive, were limited to the second-best-performing statistical group, highlighting the fact that preserving local neighborhood information through kNN connectivity and attention-based learning of weight was more effective than introducing hierarchical pooling (Fig.~\ref{Fig:Tukey_HSD}). Overall, these results support three main conclusions: kNN connectivity was consistently more effective than ASG connectivity in the attention-based models, Graph U-Net did not improve over the standard GAT-kNN framework despite producing competitive results, and GAT-kNN-\(\mathcal{L}_{\mathrm{WCE}}\) provided the best tradeoff between high mean performance and run-to-run stability. Therefore, GAT-kNN-\(\mathcal{L}_{\mathrm{WCE}}\) is selected as the final \textit{SEAGAN} model.

\subsubsection{Implications for automated A--Ci analysis}

These results have direct implications for automated A--Ci curve analysis, where the central difficulty is assigning each measurement point to the correct biochemical limitation regime before or during parameter estimation \citep{Long2003,Sharkey2007,Gu2010}. Tests on synthetically generated A--Ci curves showed that the proposed SEAGAN model substantially outperformed the automated-fitting-based PhoTorch benchmark (Table~\ref{Tab:model_summary}). PhoTorch \citep{Lei2025} achieved an F1-score of $0.735$, recall of $0.753$, precision of $0.751$, and accuracy of $0.751$. In comparison, SEAGAN achieved an F1-score of $0.857$, recall of $0.861$, precision of $0.853$, and accuracy of $0.882$. Thus, SEAGAN provides a more accurate and balanced identification of limiting states than the existing automated fitting-based benchmark. Although it can be argued that the main functionality of the PhoTorch model is not limiting state identification along the A--Ci curve, this step remains the most crucial in photosynthetic parameter optimization. Furthermore, since the PhoTorch model performs FvCB parameter optimization, the pointwise limitations identified post hoc using the estimated parameters should still match the correct limitations; only then can the parameter estimates be relied upon. The qualitative comparison in \figpanels{Fig:GNN_Edge_Ablation}{a} further supports this result. PhoTorch largely captures the pointwise progression from the \(A_c\)- to \(A_j\)- and \(A_p\)-limited regions, but incorrectly identifies the \(A_c \rightarrow A_j\) transition. In contrast, SEAGAN follows the true sequence more closely.

In practical terms, these results suggest that graph-based limitation-state classification can serve as a useful pre-fitting step for automated estimation of photosynthetic parameters. Rather than relying solely on numerical optimization to infer both limiting states and biochemical parameters simultaneously \citep{Lei2025,Lochocki2025}, the proposed approach first identifies the likely limitation regime at each measurement point. A more reliable identification of limiting regimes can support a stable estimation of photosynthetic parameters such as \(V_\text{cmax}\), \(J\), TPU, \(g_\text{m}\), and \(R_\text{d}\). Although downstream parameter optimization is not performed in this study, the improved accuracy over PhoTorch indicates that SEAGAN provides a foundation for building a scalable pipeline to analyze noisy A--Ci curves using a robust graph-based framework.

\subsubsection{Limitations and future directions}

The present study focused on identifying the active photosynthetic limitation state at each point along the CO$_2$ response curve, which describes how the leaf carbon uptake rate changes with the leaf’s internal CO$_2$ concentration. This study is limited to only the limitation state identification step and does not perform the downstream photosynthetic parameter estimation. This was a deliberate scope choice, as the major focus of this work is to identify whether graph-based learning can identify regions of the curve corresponding to distinct limiting processes. Although \textit{SEAGAN} was the best-performing model overall, some uncertainty remained near transition regions. In particular, none of the models discussed here successfully identified the second transition point (\(A_j \rightarrow A_p\)) into the TPU-limited region for the representative curve shown in \figpanels{Fig:GNN_Edge_Ablation}{a}.

Future work should connect this classification step to the estimation of photosynthetic parameters. In such a workflow, \textit{SEAGAN}'s pointwise limitation-state classification will guide the piecewise optimization of the FvCB parameters. This would allow a more stable estimation of the photosynthetic parameters (\(V_{\mathrm{cmax}}\), \(J\), TPU, \(g_m\), and \(R_d\)). An additional extension would be to develop a graph-to-parameter learning framework in which the CO$_2$ response curve is provided as a graph, and the model directly predicts photosynthetic parameters along with their associated uncertainties, without the need for the intermediate piecewise optimization step. In such a framework, graph connections will be used not only for message passing but also to represent uncertainty and dependency among neighboring points. Thus, the present study should be considered a first step toward more automated and reliable interpretation of CO$_2$ response curves.

\section{Conclusions}

The objective of this study was to investigate whether the limiting-state identification step in the photosynthetic parameter estimation process using the photosynthetic CO$_2$ response curve can be reliably estimated from graph-based analysis of the curve. To achieve this, the paper compared a combination of models ranging from classical ML-based models to deep learning models, different variants of graph-based models, and an existing automated parameter estimation tool. The results show that explicitly representing relationships among measurement points as a graph improves the limitation-state classification relative to the classical machine learning and deep learning baselines, as well as the best existing automated approach. Across the models tested in this study, the best-performing combination included local kNN-based connections, attention-based graph learning, and a weighted cross-entropy loss (GAT-kNN-\(\mathcal{L}_{\mathrm{WCE}}\)). This shows that adaptive weighting of neighboring points in the observed response space and proximity-based graph connections are most effective for identifying biochemical limitation states. \textit{SEAGAN} also outperformed the automated fitting-based tool (PhoTorch), demonstrating its advantage over the direct photosynthetic parameter estimation tools. GAT-kNN-\(\mathcal{L}_{\mathrm{WCE}}\), referred to as \textit{SEAGAN} (\textit{domain-Specific and Edge-Aware Graph Attention Network for Dynamic Plant Processes}), achieved an F1-score, recall, precision, and accuracy of $0.857$, $0.861$, $0.853$, and $0.882$, respectively. Hierarchical pooling improved the GCN-based models, but was ineffective in GAT-based models. The graph-based learning framework discussed in this study is a first step towards a functional plant phenomics workflow for interpreting leaf-level physiological responses beyond static trait extraction. Future integration with photosynthetic parameter estimates will provide a reliable tool for automated extraction of functional features from leaf gas-exchange data.

\section*{CRediT authorship contribution statement}
\textbf{Antriksh Srivastava:} Conceptualization, Methodology, Software, Validation, Formal analysis, Investigation, Data curation, Visualization, Writing -- original draft, Writing -- review and editing. \textbf{Soumyashree Kar:} Conceptualization, Methodology, Supervision, Project administration, Critical review, Writing -- review and editing.

\section*{Declaration of competing interest}
The authors declare that they have no known competing financial interests or personal relationships that could have appeared to influence the work reported in this paper.


\footnotesize
\bibliographystyle{elsarticle-num}
\bibliography{Reference}

@Article{Sharkey2007,
  author   = {Sharkey, Thomas D. and Bernacchi, Carl J. and Farquhar, Graham D. and Singsaas, Eric L.},
  journal  = {Plant, Cell \& Environment},
  title    = {Fitting photosynthetic carbon dioxide response curves for {C3} leaves},
  year     = {2007},
  number   = {9},
  pages    = {1035--1040},
  volume   = {30},
  doi      = {10.1111/j.1365-3040.2007.01710.x},
}

@Article{Duursma2015,
  author    = {Duursma, Remko A.},
  journal   = {PLOS ONE},
  title     = {{Plantecophys - An R Package for Analysing and Modelling Leaf Gas Exchange Data}},
  year      = {2015},
  month     = {11},
  number    = {11},
  pages     = {1--13},
  volume    = {10},
  doi       = {10.1371/journal.pone.0143346},
  publisher = {Public Library of Science},
}

@Article{Lei2025,
  author  = {Lei, Tong and Rizzo, Kyle T. and Bailey, Brian N.},
  journal = {Photosynthesis Research},
  title   = {{PhoTorch: a robust and generalized biochemical photosynthesis model fitting package based on PyTorch}},
  year    = {2025},
  issn    = {1573-5079},
  number  = {21},
  volume  = {163},
  doi     = {10.1007/s11120-025-01136-7},
}

@Article{Lochocki2025,
  author   = {Lochocki, Edward B. and Salesse-Smith, Coralie E. and McGrath, Justin M.},
  journal  = {Plant, Cell \& Environment},
  title    = {{PhotoGEA: An R Package for Closer Fitting of Photosynthetic Gas Exchange Data With Non-Gaussian Confidence Interval Estimation}},
  year     = {2025},
  number   = {7},
  pages    = {5104--5119},
  volume   = {48},
  doi      = {10.1111/pce.15501},
}

@inproceedings{Petar2018,
  title     = {Graph Attention Networks},
  author    = {Veli{\v{c}}kovi{\'c}, Petar and Cucurull, Guillem and Casanova, Arantxa and Romero, Adriana and Li{\`o}, Pietro and Bengio, Yoshua},
  booktitle = {Proceedings of the 6th International Conference on Learning Representations (ICLR 2018)},
  year      = {2018},
  address   = {Vancouver, BC, Canada},
  publisher = {OpenReview.net},
  note      = {Conference Track Proceedings, April 30--May 3, 2018},
  url       = {https://openreview.net/forum?id=rJXMpikCZ}
}

@article{Farquhar1980,
    title     = {{A biochemical model of photosynthetic CO$_2$ assimilation in leaves of C$_3$ species}},
    volume    = {149},
    ISSN      = {1432-2048},
    DOI       = {10.1007/bf00386231},
    number    = {1},
    journal   = {Planta},
    publisher = {Springer Science and Business Media LLC},
    author    = {Farquhar, G. D. and von Caemmerer, S. and Berry, J. A.},
    year      = {1980},
    pages     = {78–90},
}

@Book{vonCaemmerer2000,
  author    = {von Caemmerer, Susanne},
  publisher = {CSIRO},
  title     = {{Biochemical models of leaf photosynthesis}},
  year      = {2000},
  isbn      = {0-643-06379-X},
  series    = {{Techniques in plant sciences}},
  doi       = {10.1071/9780643103405},
  pages     = {xxi+286p},
  type      = {Book},
}

@article{Mckay1979,
  author    = {McKay, M. D. and Beckman, R. J. and Conover, W. J.},
  title     = {A Comparison of Three Methods for Selecting Values of Input Variables in the Analysis of Output from a Computer Code},
  journal   = {Technometrics},
  volume    = {21},
  number    = {2},
  pages     = {239--245},
  year      = {1979},
  doi       = {10.2307/1268522},
}

@Article{Moualeu2017,
  author   = {Moualeu-Ngangue, Dany P. and Chen, Tsu-Wei and St{\"u}tzel, Hartmut},
  journal  = {New Phytologist},
  title    = {A new method to estimate photosynthetic parameters through net assimilation rate-intercellular space {Co$_2$} concentration {A-Ci} curve and chlorophyll fluorescence measurements},
  year     = {2017},
  number   = {3},
  pages    = {1543--1554},
  volume   = {213},
  doi      = {10.1111/nph.14260}
}

@article{Bernacchi2001,
author = {Bernacchi, C. J. and Singsaas, E. L. and Pimentel, C. and Portis Jr, A. R. and Long, S. P.},
title = {Improved temperature response functions for models of Rubisco-limited photosynthesis},
journal = {Plant, Cell \& Environment},
volume = {24},
number = {2},
pages = {253-259},
doi = {10.1111/j.1365-3040.2001.00668.x},
year = {2001}
}

@article{Bernacchi2002,
    author = {Bernacchi, Carl J. and Portis, Archie R. and Nakano, Hiromi and von Caemmerer, Susanne and Long, Stephen P.},
    title = {Temperature Response of Mesophyll Conductance. Implications for the Determination of Rubisco Enzyme Kinetics and for Limitations to Photosynthesis in Vivo},
    journal = {Plant Physiology},
    volume = {130},
    number = {4},
    pages = {1992-1998},
    year = {2002},
    month = {12},
    issn = {0032-0889},
    doi = {10.1104/pp.008250},
}

@article{Bernacchi2003,
author = {Bernacchi, Carl J. and Pimentel, C. and Long, S. P.},
title = {{In vivo temperature response functions of parameters required to model RuBP-limited photosynthesis}},
journal = {Plant, Cell \& Environment},
volume = {26},
number = {9},
pages = {1419-1430},
doi = {10.1046/j.0016-8025.2003.01050.x},
year = {2003}
}

@article{Harley1992,
author = {Harley, P. C. and Thomas, R. B. and Reynolds, J. F. and Strain, B. R.},
title = {{Modelling photosynthesis of cotton grown in elevated CO$_2$}},
journal = {Plant, Cell \& Environment},
volume = {15},
number = {3},
pages = {271-282},
doi = {10.1111/j.1365-3040.1992.tb00974.x},
year = {1992}
}

@article{Kar2024,
title = {{XWaveNet}: Enabling uncertainty quantification in short-term ocean wave height forecasts and extreme event prediction},
journal = {Applied Ocean Research},
volume = {148},
pages = {103994},
year = {2024},
issn = {0141-1187},
doi = {10.1016/j.apor.2024.103994},
author = {Soumyashree Kar and Jason R. McKenna and Vishwamithra Sunkara and Robert Coniglione and Steve Stanic and Landry Bernard},
}

@article{Srivastava2014,
  author  = {Nitish Srivastava and Geoffrey Hinton and Alex Krizhevsky and Ilya Sutskever and Ruslan Salakhutdinov},
  title   = {Dropout: A Simple Way to Prevent Neural Networks from Overfitting},
  journal = {Journal of Machine Learning Research},
  year    = {2014},
  volume  = {15},
  number  = {56},
  pages   = {1929-1958},
  url     = {http://jmlr.org/papers/v15/srivastava14a.html}
  }

@InProceedings{Gao2019,
  title = 	 {{Graph U-Nets}},
  author =       {Gao, Hongyang and Ji, Shuiwang},
  booktitle = 	 {Proceedings of the 36th International Conference on Machine Learning},
  pages = 	 {2083--2092},
  year = 	 {2019},
  editor = 	 {Chaudhuri, Kamalika and Salakhutdinov, Ruslan},
  volume = 	 {97},
  series = 	 {Proceedings of Machine Learning Research},
  month = 	 {09--15 Jun},
  publisher =    {PMLR},
  url = 	 {https://proceedings.mlr.press/v97/gao19a.html},
}

@ARTICLE{Wu2021,
  author={Wu, Zonghan and Pan, Shirui and Chen, Fengwen and Long, Guodong and Zhang, Chengqi and Yu, Philip S.},
  journal={IEEE Transactions on Neural Networks and Learning Systems}, 
  title={A Comprehensive Survey on Graph Neural Networks}, 
  year={2021},
  volume={32},
  number={1},
  pages={4-24},
  doi={10.1109/TNNLS.2020.2978386}}

@article{Zhou2020,
title = {Graph neural networks: A review of methods and applications},
journal = {AI Open},
volume = {1},
pages = {57-81},
year = {2020},
issn = {2666-6510},
doi = {10.1016/j.aiopen.2021.01.001},
author = {Jie Zhou and Ganqu Cui and Shengding Hu and Zhengyan Zhang and Cheng Yang and Zhiyuan Liu and Lifeng Wang and Changcheng Li and Maosong Sun}
}

@article{Bronstein2021,
      title="{Geometric Deep Learning: Grids, Groups, Graphs, Geodesics, and Gauges}", 
      author={Michael M. Bronstein and Joan Bruna and Taco Cohen and Petar Veličković},
      year={2021},
      journal={arXiv preprint},
      archiveprefix = {arXiv},
      doi           = {10.48550/arXiv.2104.13478}, 	
      primaryclass  = {cs.LG},      
}

@ARTICLE{Franco2009,
  author={Scarselli, Franco and Gori, Marco and Tsoi, Ah Chung and Hagenbuchner, Markus and Monfardini, Gabriele},
  journal={IEEE Transactions on Neural Networks}, 
  title={The Graph Neural Network Model}, 
  year={2009},
  volume={20},
  number={1},
  pages={61-80},
  doi={10.1109/TNN.2008.2005605}}

@misc{Battaglia2018,
      title={Relational inductive biases, deep learning, and graph networks}, 
      author={Peter W. Battaglia and Jessica B. Hamrick and Victor Bapst and Alvaro Sanchez-Gonzalez and Vinicius Zambaldi and Mateusz Malinowski and Andrea Tacchetti and David Raposo and Adam Santoro and Ryan Faulkner and Caglar Gulcehre and Francis Song and Andrew Ballard and Justin Gilmer and George Dahl and Ashish Vaswani and Kelsey Allen and Charles Nash and Victoria Langston and Chris Dyer and Nicolas Heess and Daan Wierstra and Pushmeet Kohli and Matt Botvinick and Oriol Vinyals and Yujia Li and Razvan Pascanu},
      year={2018},
      journal={arXiv preprint},
      archivePrefix={arXiv},
      primaryClass={cs.LG},
      doi={10.48550/arXiv.2104.13478}
}

@article{Bonan2014,
   author = {G. B. Bonan and M. Williams and R. A. Fisher and K. W. Oleson},
   doi = {10.5194/gmd-7-2193-2014},
   issn = {19919603},
   issue = {5},
   journal = {Geoscientific Model Development},
   month = {9},
   pages = {2193-2222},
   publisher = {Copernicus GmbH},
   title = {Modeling stomatal conductance in the earth system: Linking leaf water-use efficiency and water transport along the soil-plant-atmosphere continuum},
   volume = {7},
   year = {2014},
}

@article{Lochocki2022,
    author = {Lochocki, Edward B and Rohde, Scott and Jaiswal, Deepak and Matthews, Megan L and Miguez, Fernando and Long, Stephen P and McGrath, Justin M},
    title = {{BioCro II: a software package for modular crop growth simulations}},
    journal = {in silico Plants},
    volume = {4},
    number = {1},
    pages = {diac003},
    year = {2022},
    month = {01},
    issn = {2517-5025},
    doi = {10.1093/insilicoplants/diac003},
}

@article{LeCun2015,
  author  = {LeCun, Yann and Bengio, Yoshua and Hinton, Geoffrey},
  title   = {Deep Learning},
  journal = {Nature},
  year    = {2015},
  volume  = {521},
  number  = {7553},
  pages   = {436--444},
  month   = {05},
  doi     = {10.1038/nature14539},
  issn    = {1476-4687},
}

@inproceedings{Thomas2017,
  author       = {Kipf, Thomas N. and Welling, Max},
  title        = {Semi-Supervised Classification with Graph Convolutional Networks},
  booktitle    = {5th International Conference on Learning Representations, {ICLR} 2017},
  year         = {2017},
  url          = {https://openreview.net/forum?id=SJU4ayYgl},
}

@inproceedings{William2017,
author = {Hamilton, William L. and Ying, Rex and Leskovec, Jure},
title = {Inductive representation learning on large graphs},
year = {2017},
isbn = {9781510860964},
publisher = {Curran Associates Inc.},
address = {Red Hook, NY, USA},
booktitle = {Proceedings of the 31st International Conference on Neural Information Processing Systems},
pages = {1025–1035},
numpages = {11},
location = {Long Beach, California, USA},
series = {NIPS'17},
url={https://papers.nips.cc/paper_files/paper/2017/hash/5dd9db5e033da9c6fb5ba83c7a7ebea9-Abstract.html}
}

@article{George2010,
author = {Forman, George and Scholz, Martin},
title = {Apples-to-apples in cross-validation studies: pitfalls in classifier performance measurement},
year = {2010},
issue_date = {June 2010},
publisher = {Association for Computing Machinery},
address = {New York, NY, USA},
volume = {12},
number = {1},
issn = {1931-0145},
doi = {10.1145/1882471.1882479},
journal = {SIGKDD Explor. Newsl.},
month = nov,
pages = {49–57},
numpages = {9}
}

@article{Tukey1949,
 ISSN = {0006341X, 15410420},
 author = {John W. Tukey},
 journal = {Biometrics},
 number = {2},
 pages = {99--114},
 publisher = {[Wiley, International Biometric Society]},
 title = {Comparing Individual Means in the Analysis of Variance},
 urldate = {2026-04-30},
 volume = {5},
 year = {1949},
 doi={10.2307/3001913}
}

@article{Kar2020,
  author  = {Kar, Soumyashree and Tanaka, Ryokei and Korbu, Lijalem Balcha and Kholov{\'a}, Jana and Iwata, Hiroyoshi and Durbha, Surya S. and Adinarayana, J. and Vadez, Vincent},
  title   = {Automated discretization of `transpiration restriction to increasing VPD' features from outdoors high-throughput phenotyping data},
  journal = {Plant Methods},
  year    = {2020},
  volume  = {16},
  number  = {1},
  pages   = {140},
  month   = oct,
  doi     = {10.1186/s13007-020-00680-8},
  issn    = {1746-4811},
}

@inproceedings{Ying2019,
  author    = {Ying, Rex and Bourgeois, Dylan and You, Jiaxuan and Zitnik, Marinka and Leskovec, Jure},
  title     = {{GNNExplainer}: Generating Explanations for Graph Neural Networks},
  booktitle = {Advances in Neural Information Processing Systems},
  volume    = {32},
  pages     = {9240--9251},
  year      = {2019},
  url={https://papers.nips.cc/paper_files/paper/2019/hash/d80b7040b773199015de6d3b4293c8ff-Abstract.html}
}

@book{Fisher1992,
  author    = {Fisher, Ronald A.},
  title     = {Statistical Methods for Research Workers},
  booktitle = {Breakthroughs in Statistics: Methodology and Distribution},
  publisher = {Springer},
  address   = {New York},
  pages     = {66--70},
  year      = {1992},
  doi       = {10.1007/978-1-4612-4380-9_6},
  isbn      = {978-1-4612-4380-9},
}

@article{Long2003,
    author = {Long, S. P. and Bernacchi, C. J.},
    title = {Gas exchange measurements, what can they tell us about the underlying limitations to photosynthesis? Procedures and sources of error},
    journal = {Journal of Experimental Botany},
    volume = {54},
    number = {392},
    pages = {2393-2401},
    year = {2003},
    month = {11},
    issn = {0022-0957},
    doi = {10.1093/jxb/erg262},
}

@article{Xu2018,
    author = {Xu, Keyulu and Hu, Weihua and Leskovec, Jure and Jegelka, Stefanie},
    title = {{How Powerful are Graph Neural Networks?}},
    eprint = {1810.00826},
    archivePrefix = {arXiv},
    primaryClass = {cs.LG},
    year = {2018},
    journal={arXiv preprint},
    doi={10.48550/arXiv.1810.00826}
}

@INPROCEEDINGS{Charilaos2014,
  author={Kanatsoulis, Charilaos I. and Ribeiro, Alejandro},
  booktitle={ICASSP 2024 - 2024 IEEE International Conference on Acoustics, Speech and Signal Processing (ICASSP)}, 
  title={Graph Neural Networks are More Powerful than We Think}, 
  year={2024},
  pages={7550-7554},
  doi={10.1109/ICASSP48485.2024.10447704}
  }

@inproceedings{Sanchez2020,
author = {Sanchez-Gonzalez, Alvaro and Godwin, Jonathan and Pfaff, Tobias and Ying, Rex and Leskovec, Jure and Battaglia, Peter W.},
title = {Learning to simulate complex physics with graph networks},
year = {2020},
publisher = {JMLR.org},
booktitle = {Proceedings of the 37th International Conference on Machine Learning},
articleno = {784},
numpages = {10},
series = {ICML'20},
url = 	 {https://proceedings.mlr.press/v119/sanchez-gonzalez20a.html}
}

@article{Furbank2011,
  author    = {Robert T. Furbank and Mark Tester},
  title     = {Phenomics--technologies to relieve the phenotyping bottleneck},
  journal   = {Trends in Plant Science},
  year      = {2011},
  volume    = {16},
  number    = {12},
  pages     = {635--644},
  doi       = {10.1016/j.tplants.2011.09.005},
  pmid      = {22074787},
  issn      = {1360-1385},
  month     = dec
}

@article{Dubois2007,
author = {Dubois, Jean-Jacques B. and Fiscus, Edwin L. and Booker, Fitzgerald L. and Flowers, Michael D. and Reid, Chantal D.},
title = {Optimizing the statistical estimation of the parameters of the Farquhar–von Caemmerer–Berry model of photosynthesis},
journal = {New Phytologist},
volume = {176},
number = {2},
pages = {402-414},
doi = {10.1111/j.1469-8137.2007.02182.x},
year = {2007}
}

@article{Gu2010,
author = {Gu, Lianhong and Pallardy, Stephen G. and Tu, Kevin and Law, Beverly E. and Wullschleger, Stan D.},
title = {Reliable estimation of biochemical parameters from C3 leaf photosynthesis–intercellular carbon dioxide response curves},
journal = {Plant, Cell \& Environment},
volume = {33},
number = {11},
pages = {1852-1874},
doi = {10.1111/j.1365-3040.2010.02192.x},
year = {2010}
}

@inproceedings{Chen2016,
author = {Chen, Tianqi and Guestrin, Carlos},
title = {{XGBoost: A Scalable Tree Boosting System}},
year = {2016},
isbn = {9781450342322},
publisher = {Association for Computing Machinery},
address = {New York, NY, USA},
doi = {10.1145/2939672.2939785},
booktitle = {Proceedings of the 22nd ACM SIGKDD International Conference on Knowledge Discovery and Data Mining},
pages = {785–794},
numpages = {10},
location = {San Francisco, California, USA},
series = {KDD '16}
}

@inproceedings{Lin2017,
  author    = {Tsung{-}Yi Lin and Priya Goyal and Ross Girshick and Kaiming He and Piotr Doll{\'a}r},
  title     = {Focal Loss for Dense Object Detection},
  booktitle = {Proceedings of the IEEE International Conference on Computer Vision (ICCV)},
  pages     = {2999--3007},
  year      = {2017},
  doi       = {10.1109/ICCV.2017.324}
}

@article{Cortes1995,
  author    = {Corinna Cortes and Vladimir Vapnik},
  title     = {Support-vector networks},
  journal   = {Machine Learning},
  year      = {1995},
  volume    = {20},
  number    = {3},
  pages     = {273--297},
  doi       = {10.1007/BF00994018},
  issn      = {1573-0565},
}

@article{Breiman2001,
  author    = {Leo Breiman},
  title     = {Random Forests},
  journal   = {Machine Learning},
  year      = {2001},
  volume    = {45},
  number    = {1},
  pages     = {5--32},
  doi       = {10.1023/A:1010933404324},
  issn      = {1573-0565},
}

@article{He2009,
  author={He, Haibo and Garcia, Edwardo A.},
  journal={IEEE Transactions on Knowledge and Data Engineering}, 
  title={Learning from Imbalanced Data}, 
  year={2009},
  volume={21},
  number={9},
  pages={1263-1284},
  doi={10.1109/TKDE.2008.239}}

@inproceedings{Elkan2001,
author = {Elkan, Charles},
title = {The foundations of cost-sensitive learning},
year = {2001},
isbn = {1558608125},
publisher = {Morgan Kaufmann Publishers Inc.},
address = {San Francisco, CA, USA},
booktitle = {Proceedings of the 17th International Joint Conference on Artificial Intelligence - Volume 2},
pages = {973–978},
numpages = {6},
location = {Seattle, WA, USA},
series = {IJCAI'01},
url={https://dl.acm.org/doi/10.5555/1642194.1642224}
}

@article{You2017, 
author={You, Jiaxuan and Li, Xiaocheng and Low, Melvin and Lobell, David and Ermon, Stefano}, 
title={Deep Gaussian Process for Crop Yield Prediction Based on Remote Sensing Data}, 
volume={31},
DOI={10.1609/aaai.v31i1.11172},
number={1}, 
journal={Proceedings of the AAAI Conference on Artificial Intelligence}, 
year={2017}}

@Article{Khaki2019,
  author  = {Khaki, Saeed and Wang, Lizhi},
  journal = {Frontiers in Plant Science},
  title   = {Crop Yield Prediction Using Deep Neural Networks},
  year    = {2019},
  issn    = {1664-462X},
  volume  = {10},
  doi     = {10.3389/fpls.2019.00621},
}

@article{Nevavuori2019,
title = {Crop yield prediction with deep convolutional neural networks},
author = {Petteri Nevavuori and Nathaniel Narra and Tarmo Lipping},
journal = {Computers and Electronics in Agriculture},
volume = {163},
pages = {104859},
year = {2019},
issn = {0168-1699},
doi = {10.1016/j.compag.2019.104859},
}

@article{Riera2021,
title = {Deep Multiview Image Fusion for Soybean Yield Estimation in Breeding Applications},
journal = {Plant Phenomics},
volume = {2021},
pages = {9846470},
year = {2021},
issn = {2643-6515},
doi = {10.34133/2021/9846470},
author = {Luis G. Riera and Matthew E. Carroll and Zhisheng Zhang and Johnathon M. Shook and Sambuddha Ghosal and Tianshuang Gao and Arti Singh and Sourabh Bhattacharya and Baskar Ganapathysubramanian and Asheesh K. Singh and Soumik Sarkar},
}

@article{Shook2021,
    doi = {10.1371/journal.pone.0252402},
    author = {Shook, Johnathon AND Gangopadhyay, Tryambak AND Wu, Linjiang AND Ganapathysubramanian, Baskar AND Sarkar, Soumik AND Singh, Asheesh K.},
    journal = {PLOS ONE},
    publisher = {Public Library of Science},
    title = {Crop yield prediction integrating genotype and weather variables using deep learning},
    year = {2021},
    month = {06},
    volume = {16},
    pages = {1-19},
    number = {6},
}

@article{Li2022,
  title={{UAV-based hyperspectral and ensemble machine learning for predicting yield in winter wheat}},
  author={Li, Zongpeng and Chen, Zhen and Cheng, Qian and Duan, Fuyi and Sui, Ruixiu and Huang, Xiuqiao and Xu, Honggang},
  journal={Agronomy},
  volume={12},
  number={1},
  pages={202},
  year={2022},
  doi={10.3390/agronomy12010202},
  publisher={MDPI}
}

@article{Fan2022, 
title={{A GNN-RNN Approach for Harnessing Geospatial and Temporal Information: Application to Crop Yield Prediction}}, 
volume={36},
DOI={10.1609/aaai.v36i11.21444},
number={11}, 
journal={Proceedings of the AAAI Conference on Artificial Intelligence}, 
author={Fan, Joshua and Bai, Junwen and Li, Zhiyun and Ortiz-Bobea, Ariel and Gomes, Carla P.}, 
year={2022},
pages={11873-11881} 
}

@article{Sajitha2023,
title = {Smart farming application using knowledge embedded-graph convolutional neural network ({KEGCNN}) for banana quality detection},
journal = {Journal of Agriculture and Food Research},
volume = {14},
pages = {100767},
year = {2023},
issn = {2666-1543},
doi = {10.1016/j.jafr.2023.100767},
author = {P. Sajitha and A. {Diana Andrushia} and Nour Mostafa and Ahmed {Younes Shdefat} and S.S. Suni and N. Anand},
}

@misc{Gupta2023,
      title={Agri-GNN: A Novel Genotypic-Topological Graph Neural Network Framework Built on GraphSAGE for Optimized Yield Prediction}, 
      author={Aditya Gupta and Asheesh Singh},
      year={2023},
      journal={arXiv preprint},
      archivePrefix={arXiv},
      primaryClass={cs.LG},
      doi={10.48550/arXiv.2310.13037}, 
}

@Article{Collatz1991,
  author    = {Collatz, G. J. and Ball, J. T. and Grivet, C. and Berry, J. A.},
  journal   = {Agricultural and Forest meteorology},
  title     = {Physiological and environmental regulation of stomatal conductance, photosynthesis and transpiration: a model that includes a laminar boundary layer},
  year      = {1991},
  month     = {apr},
  number    = {2-4},
  pages     = {107--136},
  volume    = {54},
  doi       = {10.1016/0168-1923(91)90002-8},
  publisher = {Elsevier},
}

@article{Gregory2021,
author = {Gregory, Luke M. and McClain, Alan M. and Kramer, David M. and Pardo, Jeremy D. and Smith, Kaila E. and Tessmer, Oliver L. and Walker, Berkley J. and Ziccardi, Leonardo G. and Sharkey, Thomas D.},
title = {The triose phosphate utilization limitation of photosynthetic rate: Out of global models but important for leaf models},
journal = {Plant, Cell \& Environment},
volume = {44},
number = {10},
pages = {3223-3226},
doi = {10.1111/pce.14153},
year = {2021}
}

@Article{Yulong2019,
  author  = {Ma, Yulong and Liu, Heping},
  journal = {Journal of Advances in Modeling Earth Systems},
  title   = {{An Advanced Multiple-Layer Canopy Model in the WRF Model With Large-Eddy Simulations to Simulate Canopy Flows and Scalar Transport Under Different Stability Conditions}},
  year    = {2019},
  number  = {7},
  pages   = {2330--2351},
  volume  = {11},
  doi     = {10.1029/2018MS001347},
}

@Article{Srivastava2024,
  author   = {Srivastava, Antriksh and Srinivasan, Venkatraman and Long, Stephen P.},
  journal  = {Plant, Cell \& Environment},
  title    = {Stomatal conductance reduction tradeoffs in maize leaves: A theoretical study},
  year     = {2024},
  number   = {5},
  pages    = {1716--1731},
  volume   = {47},
  doi      = {10.1111/pce.14821},
}

@article{Zhou2019,
  author  = {Zhou, Haoran and Ak{\c c}ay, Erol and Helliker, Brent R.},
  title   = {Estimating {C4} photosynthesis parameters by fitting intensive {A/Ci} curves},
  journal = {Photosynthesis Research},
  year    = {2019},
  volume  = {141},
  number  = {2},
  pages   = {181--194},
  doi     = {10.1007/s11120-019-00619-8},
  issn    = {1573-5079}
}

@inproceedings{Lenssen2019,
  title={Fast Graph Representation Learning with {PyTorch Geometric}},
  author={Fey, Matthias and Lenssen, Jan E.},
  booktitle={ICLR Workshop on Representation Learning on Graphs and Manifolds},
  year={2019},
  url = {https://arxiv.org/abs/1903.02428},
}

\setcounter{section}{0}
\renewcommand{\thesection}{\Large{S\arabic{section}}}
\setcounter{figure}{0}
\renewcommand{\thefigure}{S\arabic{figure}}
\setcounter{table}{0}
\renewcommand{\thetable}{S\arabic{table}}

\clearpage

\section{\Large{Supplementary Material}}

\begin{longtable}{p{1.2cm} p{11cm} c l}
\caption{Summary of variables and notation used in the manuscript.}
\label{Tab:variables_notation}\\

\toprule
\textbf{Variable} & \textbf{Description} & \textbf{Value} & \textbf{Dimension} \\
\midrule
\endfirsthead

\toprule
\textbf{Variable} & \textbf{Description} & \textbf{Value} & \textbf{Dimension} \\
\midrule
\endhead

\bottomrule
\endfoot

\multicolumn{4}{c}{\textbf{Data generation and curve sampling}}\\
\midrule

$N$ 
& Total number of synthetic A--Ci curves generated in the dataset. 
& $10000$ 
& Dimensionless \\

$m$ 
& Number of points in a given curve after subsampling and sorting. 
& Variable 
& Dimensionless \\

$n$ 
& Number of directed edges in the graph after kNN or ASG graph construction, i.e., \(n=|\mathcal{E}|\). 
& Variable 
& Dimensionless \\

$j$ 
& Index of a measurement point or graph node. 
& Variable 
& Dimensionless  \\

$r$ 
& Rank index after sorting A--Ci points by increasing $C_\mathrm{i}$. 
& Variable 
& Dimensionless  \\

$T_\text{leaf}$ 
& Leaf temperature used during the A--Ci curve simulation. 
& Variable 
& $\si{\degreeCelsius}$ \\

$P_\text{atm}$ 
& Atmospheric pressure used during gas-exchange simulation. 
& $101.325$ 
& $\si{\kilo\pascal}$ \\

$\mathrm{O}_2$ 
& Oxygen partial pressure used during gas-exchange simulation. 
& $210$ 
& $\si{\milli\bar}$ \\

\midrule
\multicolumn{4}{c}{\textbf{Photosynthesis variables}}\\
\midrule

$C_\mathrm{i}$ 
& Intercellular CO$_2$ concentration at a measurement point. 
& Variable 
& $\si{\micro\mol\per\mol}$ \\

$C_c$ 
& Chloroplast CO$_2$ concentration. 
& Variable 
& $\si{\micro\mol\per\mol}$ \\

$A_\text{net}$ 
& Net CO$_2$ assimilation rate. 
& Variable 
& $\si{\micro\mol\per\meter\squared\per\second}$ \\

$A_c$ 
& Rubisco-limited net assimilation rate. 
& Variable 
& $\si{\micro\mol\per\meter\squared\per\second}$ \\

$A_j$ 
& RuBP-regeneration (electron-transport) -limited net assimilation rate. 
& Variable 
& $\si{\micro\mol\per\meter\squared\per\second}$ \\

$A_p$ 
& Triose-phosphate-utilization-limited net assimilation rate. 
& Variable 
& $\si{\micro\mol\per\meter\squared\per\second}$ \\

$W_c$ 
& Gross CO$_2$ assimilation rate under Rubisco limitation. 
& Variable 
& $\si{\micro\mol\per\meter\squared\per\second}$ \\

$W_j$ 
& Gross CO$_2$ assimilation rate under RuBP-regeneration limitation. 
& Variable 
& $\si{\micro\mol\per\meter\squared\per\second}$ \\

$W_p$ 
& Gross CO$_2$ assimilation rate under TPU limitation. 
& Variable 
& $\si{\micro\mol\per\meter\squared\per\second}$ \\

$V_{\text{cmax},25}$ 
& Maximum Rubisco carboxylation capacity at $25^\circ$C. 
& Variable 
& $\si{\micro\mol\per\meter\squared\per\second}$ \\

$J$ 
& Electron transport rate. 
& Variable 
& $\si{\micro\mol\per\meter\squared\per\second}$ \\

$J_{\text{max},25}$ 
& Maximum electron transport rate at $25^\circ$C. 
& Variable 
& $\si{\micro\mol\per\meter\squared\per\second}$ \\

$\mathrm{TPU}_{25}$ 
& Triose phosphate utilization rate at $25^\circ$C. 
& Variable 
& $\si{\micro\mol\per\meter\squared\per\second}$ \\

$R_{\text{d},25}$ 
& Day respiration rate at $25^\circ$C. 
& Variable 
& $\si{\micro\mol\per\meter\squared\per\second}$ \\

$g_{\text{m},25}$ 
& Mesophyll conductance at $25^\circ$C. 
& Variable 
& $\si{\mol\per\meter\squared\per\second}$ \\

$K_c$ 
& Michaelis--Menten constant of Rubisco for CO$_2$. 
& Variable 
& $\si{\micro\mol\per\mol}$ \\

$K_o$ 
& Michaelis--Menten constant of Rubisco for O$_2$. 
& Variable 
& $\si{\milli\mol\per\mol}$ \\

$O_c$ 
& Chloroplast oxygen concentration. 
& Variable 
& $\si{\milli\mol\per\mol}$ \\

$\Gamma^\ast$ 
& CO$_2$ compensation point in the absence of mitochondrial respiration. 
& Variable 
& $\si{\micro\mol\per\mol}$ \\

\midrule
\multicolumn{4}{c}{\textbf{Auxiliary signal variables}}\\
\midrule

$\phi_c(j)$ 
& Rubisco-related auxiliary response factor computed at node $j$. 
& Variable 
& Dimensionless\\

$\phi_j(j)$ 
& Electron-transport-related auxiliary response factor computed at node $j$. 
& Variable 
& Dimensionless \\

$s_{Ac}(j)$ 
& Auxiliary diagnostic signal obtained by normalizing $A_\text{net}(j)$ with $\phi_c(j)$. 
& Variable 
& $\si{\micro\mol\per\meter\squared\per\second}$ \\

$s_{Aj}(j)$ 
& Auxiliary diagnostic signal obtained by normalizing $A_\text{net}(j)$ with $\phi_j(j)$. 
& Variable 
& $\si{\micro\mol\per\meter\squared\per\second}$ \\

$\pi(\cdot)$ 
& Permutation operator used to sort the A--Ci curve by increasing $C_\mathrm{i}$. 
& Variable 
& Operator \\

$\tilde{C}_i(r)$ 
& Intercellular CO$_2$ concentration after sorting by $C_\mathrm{i}$. 
& Variable 
& $\si{\micro\mol\per\mol}$ \\

$\tilde{A}_\text{net}(r)$ 
& Net assimilation rate after sorting by $C_\mathrm{i}$. 
& Variable 
& $\si{\micro\mol\per\meter\squared\per\second}$ \\

$\tilde{s}_{Ac}(r)$ 
& Sorted Rubisco-related auxiliary signal. 
& Variable 
& Model-dependent \\

$\tilde{s}_{Aj}(r)$ 
& Sorted electron-transport-related auxiliary signal. 
& Variable 
& Model-dependent \\

$\mathcal{P}(\cdot)$ 
& Peak-detection operator applied to the sorted auxiliary signals. 
& Variable 
& Operator \\

$p_c$ 
& Peak index detected from the sorted $s_{Ac}$ signal. 
& Variable 
& Dimensionless  \\

$p_j$ 
& Peak index detected from the sorted $s_{Aj}$ signal. 
& Variable 
& Dimensionless  \\

$\mathcal{G}_0$ 
& First ASG group, containing nodes with $r \le p_c$. 
& Variable 
& Set of nodes \\

$\mathcal{G}_1$ 
& Second ASG group, containing nodes with $p_c < r \le p_j$. 
& Variable 
& Set of nodes \\

$\mathcal{G}_2$ 
& Third ASG group, containing nodes with $r > p_j$. 
& Variable 
& Set of nodes \\

\midrule
\multicolumn{4}{c}{\textbf{Graph construction variables}}\\
\midrule

$G$ 
& Graph representation of an A--Ci curve. 
& Variable 
& Graph \\

$\mathcal{V}$ 
& Set of nodes in the graph. 
& Variable 
& Set \\

$\mathcal{E}$
& Set of graph edges. 
& Variable 
& Set \\

$(u,v)$ 
& Directed edge from source node ($u$) to destination node ($v$) for an edge. 
& Variable 
& Edge \\

$\mathcal{N}_k(j)$ 
& Set of $k$ nearest neighbors of node $j$ in the kNN graph. 
& Variable 
& Set of nodes \\

$k$ 
& Number of nearest neighbors used for kNN graph construction. 
& $4$ 
& Dimensionless \\

$\mathcal{E}_{\mathrm{kNN}}$ 
& Edge set produced by the kNN graph construction. 
& Variable 
& Set of edges \\

$\mathcal{E}_{\mathrm{ASG}}$ 
& Edge set produced by auxiliary-signal-guided graph construction. 
& Variable 
& Set of edges \\

$\mathcal{E}_{\mathrm{boundary}}$ 
& Edge set connecting adjacent ASG groups. 
& Variable 
& Set of edges \\

$u_{01},v_{01}$ 
& Boundary nodes connecting groups $\mathcal{G}_0$ and $\mathcal{G}_1$. 
& Variable 
& Dimensionless  \\

$u_{12},v_{12}$ 
& Boundary nodes connecting groups $\mathcal{G}_1$ and $\mathcal{G}_2$. 
& Variable 
& Dimensionless  \\

$\mathbf{x}_j$ 
& Node feature vector for node $j$, containing $C_\mathrm{i}$, $A_\text{net}$, $s_{Ac}$, and $s_{Aj}$. 
& Variable 
& $\mathbb{R}^{4}$ \\

$\mathbf{e}_{uv}$ 
& Edge attribute vector for the directed edge $(u,v)$. 
& Variable 
& $\mathbb{R}^{2}$ \\

$e_{c,uv}$ 
& Edge attribute based on the difference in $s_{Ac}$ between nodes $u$ and $v$. 
& Variable 
& Model-dependent \\

$e_{j,uv}$ 
& Edge attribute based on the difference in $s_{Aj}$ between nodes $u$ and $v$. 
& Variable 
& Model-dependent \\

\midrule
\multicolumn{4}{c}{\textbf{Graph neural network variables}}\\
\midrule

$\mathbf{h}_j^{(\ell)}$ 
& Hidden representation of node $j$ at graph layer $\ell$. 
& Variable 
& Model-dependent \\

$\mathbf{H}^{(\ell)}$ 
& Matrix of hidden node representations at graph layer $\ell$. 
& Variable 
& Model-dependent \\

$\ell$ 
& Graph neural network layer index. 
& Variable 
& Dimensionless  \\

$L$ 
& Number of graph neural network layers or final layer index. 
& Variable 
& Dimensionless \\

$\hat{\mathbf{A}}$ 
& Normalized graph connectivity matrix. 
& Variable 
& $\mathbb{R}^{m \times m}$ \\

$\mathbf{W}^{(\ell)}$ 
& Learnable weight matrix at graph layer $\ell$. 
& Variable 
& Model-dependent \\

$\sigma(\cdot)$ 
& Nonlinear activation function. 
& Variable 
& Function \\

$\mathbf{Z}$ 
& Logit matrix containing class scores for all nodes. 
& Variable 
& $\mathbb{R}^{m \times 3}$ \\

$Z_{j,c}$ 
& Logit score for node $j$ and class $c$. 
& Variable 
& Dimensionless  \\

$\mathbf{W}_{\mathrm{out}}$ 
& Learnable output-layer weight matrix. 
& Variable 
& Model-dependent \\

$\mathbf{b}_{\mathrm{out}}$ 
& Learnable output-layer bias vector. 
& Variable 
& Model-dependent \\

$\hat{y}_j$ 
& Predicted class label for node $j$. 
& Variable 
& Dimensionless  \\

$\mathcal{N}(i)$ 
& Neighborhood of node $i$ used in graph message passing. 
& Variable 
& Model-dependent \\

$\alpha_{ij}^{(\ell,k)}$ 
& Attention coefficient from node $j$ to node $i$ at layer $\ell$ and attention head $k$. 
& Variable 
& Dimensionless \\

$\alpha_{ij}^{(L-1)}$ 
& Attention coefficient in the final GAT layer. 
& Variable 
& Dimensionless \\

$K$ 
& Number of attention heads in the GAT layer. 
& $5$ 
& Dimensionless \\

$p$ 
& Dropout probability during model training. 
& Variable 
& Dimensionless \\

\midrule
\multicolumn{4}{c}{\textbf{Loss functions and class labels}}\\
\midrule

$\mathcal{L}_{\mathrm{CE}}$ 
& Standard multiclass cross-entropy loss. 
& Variable 
& Dimensionless \\

$\mathcal{L}_{\mathrm{WCE}}$ 
& Weighted cross-entropy loss. 
& Variable 
& Dimensionless \\

$\mathcal{L}_{\mathrm{WF}}$ 
& Weighted focal loss. 
& Variable 
& Dimensionless \\

$y_{jc}$ 
& Binary indicator denoting whether node $j$ belongs to class $c$. 
& $0$ or $1$ 
& Dimensionless \\

$\hat{y}_{jc}$ 
& Predicted probability that node $j$ belongs to class $c$. 
& Variable 
& Dimensionless \\

$c$ 
& Class index. The classes correspond to $A_c$, $A_j$, and $A_p$ limitation. 
& Variable 
& Dimensionless  \\

$C$ 
& Number of limitation-state classes. 
& $3$ 
& Dimensionless \\

$w_c$ 
& Class-specific weight used loss function. 
& Variable 
& Dimensionless \\

$\gamma$ 
& Focusing parameter in weighted focal loss. 
& Variable 
& Dimensionless \\

\midrule
\multicolumn{4}{c}{\textbf{Evaluation metrics and statistical variables}}\\
\midrule

$TP$ 
& Number of true positives. 
& Variable 
& Dimensionless \\

$TN$ 
& Number of true negatives. 
& Variable 
& Dimensionless \\

$FP$ 
& Number of false positives. 
& Variable 
& Dimensionless \\

$FN$ 
& Number of false negatives. 
& Variable 
& Dimensionless \\

$\mathrm{Accuracy}$ 
& Proportion of correctly classified nodes. 
& Variable 
& Dimensionless \\

$\mathrm{Precision}$ 
& Proportion of predicted positive labels that are correct. 
& Variable 
& Dimensionless \\

$\mathrm{Recall}$ 
& Proportion of true positive labels that are correctly detected. 
& Variable 
& Dimensionless \\

$\mathrm{F1}$ 
& Harmonic mean of precision and recall. 
& Variable 
& Dimensionless \\

$P(A_c)$ 
& Predicted probability that a node belongs to the $A_c$-limited class. 
& Variable 
& Dimensionless \\

$P(A_j)$ 
& Predicted probability that a node belongs to the $A_j$-limited class. 
& Variable 
& Dimensionless \\

$P(A_p)$ 
& Predicted probability that a node belongs to the $A_p$-limited class. 
& Variable 
& Dimensionless \\

\midrule
\multicolumn{4}{c}{\textbf{Machine-learning and model hyperparameters}}\\
\midrule

$n_{\mathrm{est}}$ 
& Number of estimators or trees in Random Forest and XGBoost models. 
& Variable 
& Dimensionless \\

$C_{\mathrm{SVM}}$ 
& Regularization parameter in the support vector machine model. 
& Variable 
& Dimensionless \\

$\Gamma_{\mathrm{SVM}}$ 
& Kernel coefficient in the support vector machine model. 
& Variable 
& Dimensionless \\

Deg. 
& Polynomial degree used in the support vector machine model. 
& Variable 
& Dimensionless \\

LR 
& Learning rate used in XGBoost or neural-network optimization. 
& Variable 
& Dimensionless \\

Sub 
& Subsample ratio used in XGBoost. 
& Variable 
& Dimensionless \\

Col 
& Column-sampling ratio used in XGBoost. 
& Variable 
& Dimensionless \\

MCW 
& Minimum child weight used in XGBoost. 
& Variable 
& Dimensionless \\

$\alpha_{\mathrm{XGB}}$ 
& L1 regularization parameter used in XGBoost. 
& Variable 
& Dimensionless \\

$\lambda_{\mathrm{XGB}}$ 
& L2 regularization parameter used in XGBoost. 
& Variable 
& Dimensionless \\

Depth 
& Maximum tree depth or neural-network depth, depending on the model. 
& Variable 
& Dimensionless \\

Split 
& Minimum number of samples required to split an internal node in Random Forest. 
& Variable 
& Dimensionless \\

Leaf 
& Minimum number of samples required at a leaf node in Random Forest. 
& Variable 
& Dimensionless \\

Feat. 
& Number or type of features considered per split in Random Forest. 
& Variable 
& Dimensionless  \\

Layers 
& Number of hidden layers in the feed-forward neural-network baseline. 
& Variable 
& Dimensionless \\

Nodes 
& Number of neurons per hidden layer in the feed-forward neural-network baseline. 
& Variable 
& Dimensionless \\
\end{longtable}

{\fontsize{9}{9} \selectfont
\begin{table*}[h]
\centering
\caption{Best hyperparameter combinations selected for the NN, RF, SVM, and XGBoost pointwise classification models for each curve length.}
\label{Tab:ML_params}
\vspace{0.1cm}
\renewcommand{\arraystretch}{1.1}
\setlength{\tabcolsep}{1.5pt}

\begin{tabular}{c|cc|ccccc|cccc|cccccccc}
\hline
\multirow{2}{*}{\textbf{Points}} 
& \multicolumn{2}{c|}{\textbf{NN}}
& \multicolumn{5}{c|}{\textbf{Random Forest}} 
& \multicolumn{4}{c|}{\textbf{SVM}} 
& \multicolumn{8}{c}{\textbf{XGBoost}} \\
\cline{2-20}
& \textbf{Layers} & \textbf{Nodes}
& \textbf{$n_{\mathrm{est}}$} & \textbf{Depth} & \textbf{Split} & \textbf{Leaf} & \textbf{Feat.}
& \textbf{$C$} & \textbf{Kernel} & \textbf{Gamma} & \textbf{Deg.}
& \textbf{$n_{\mathrm{est}}$} & \textbf{Depth} & \textbf{LR} & \textbf{Sub} & \textbf{Col} & \textbf{MCW} & \textbf{$\alpha$} & \textbf{$\lambda$} \\
\hline
8  & 1 & 32
   & 800 & 10 & 10 & 1 & 1
   & 10   & poly & 1    & 3
   & 300 & 3 & 0.10 & 1.0 & 0.7 & 4 & 0.01 & 1.0 \\

9  & 1 & 64
   & 800 & 5  & 5  & 4 & log2
   & 0.1  & rbf  & 1    & 4
   & 100 & 6 & 0.03 & 0.8 & 0.8 & 1 & 0.01 & 1.0 \\

10 & 1 & 64
   & 400 & 5  & 5  & 2 & 4
   & 1    & rbf  & 0.1  & 3
   & 300 & 3 & 0.03 & 0.8 & 1.0 & 4 & 0.01 & 1.0 \\

11 & 1 & 64
   & 200 & 10 & 2  & 1 & 3
   & 10   & rbf  & auto & 2
   & 200 & 4 & 0.03 & 0.8 & 0.7 & 1 & 1.00 & 2.0 \\

12 & 1 & 64
   & 800 & 5  & 10 & 2 & sqrt
   & 100  & rbf  & 0.1  & 3
   & 300 & 4 & 0.03 & 0.8 & 1.0 & 2 & 0.00 & 0.5 \\

13 & 1 & 64
   & 800 & 10 & 10 & 1 & 1
   & 10   & rbf  & auto & 2
   & 300 & 8 & 0.05 & 0.8 & 0.8 & 6 & 1.00 & 5.0 \\

14 & 1 & 64
   & 200 & 10 & 10 & 2 & 1
   & 10   & rbf  & auto & 2
   & 100 & 6 & 0.03 & 0.8 & 0.8 & 1 & 0.01 & 1.0 \\

15 & 1 & 64
   & 200 & 10 & 10 & 4 & 3
   & 10   & rbf  & auto & 2
   & 200 & 6 & 0.10 & 1.0 & 0.9 & 2 & 0.01 & 1.0 \\
\hline
\end{tabular}
\end{table*}
}

\begin{figure*}[h]
    \centering
    \includegraphics[width=0.48\textwidth]{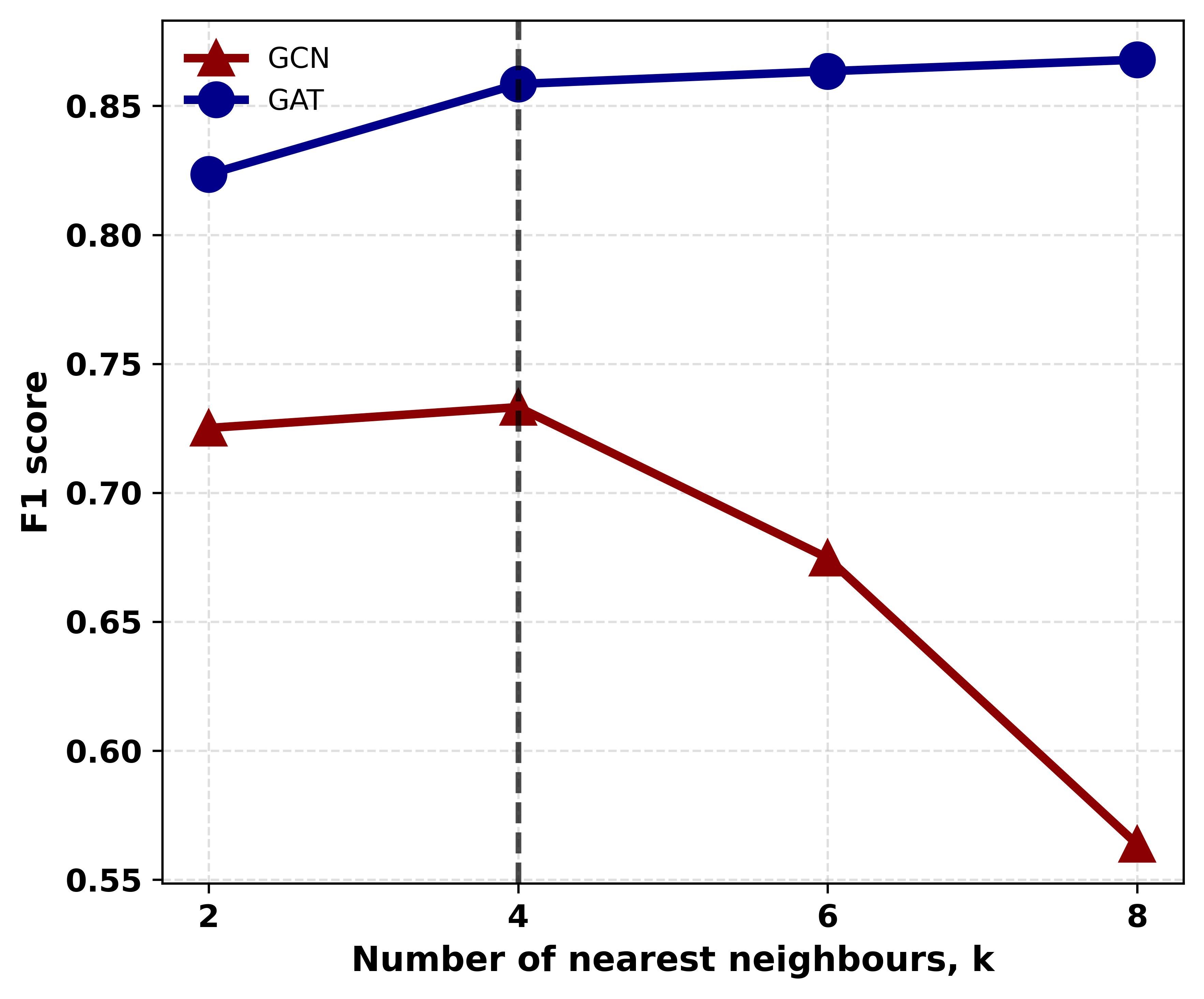}
    \caption{kNN connectivity sensitivity in graph models. Model performance was evaluated using different numbers of nearest-neighbor connections for GCN- and GAT-based graph models in the A--Ci graph.}
    \label{Fig:k_Sensitivity}
\end{figure*}

\begin{figure*}[h]
    \centering
    \includegraphics[width=0.5\textwidth]{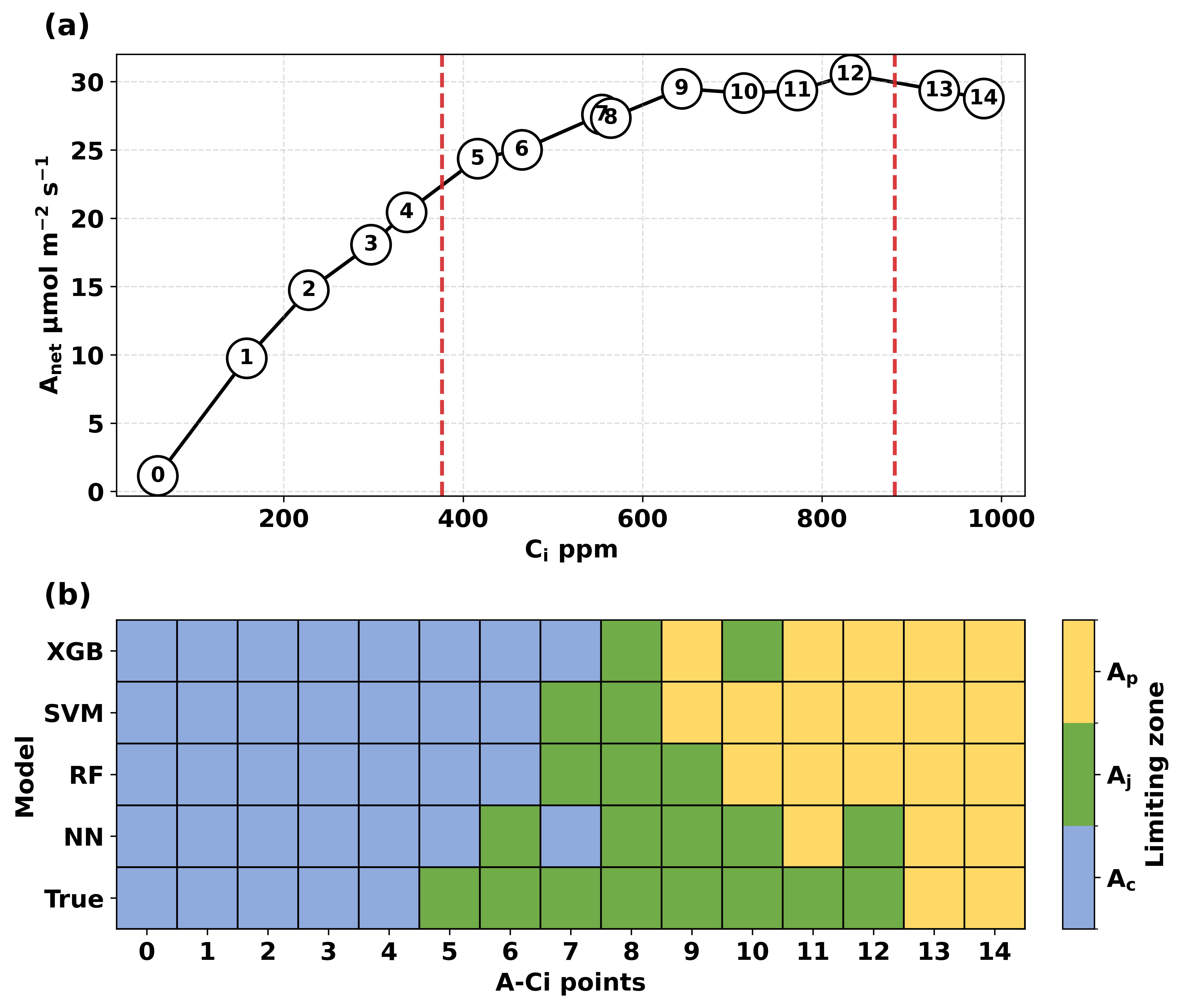}
    \caption{Feature-based ML model predictions for a representative A--Ci curve. (a) shows the measured A--Ci response with numbered nodes and dashed vertical lines marking true limitation transitions, (b) compares true and predicted limiting zones across NN, RF, SVM, and XGB models.}
    \label{Fig:ML_models_test}
\end{figure*}

\begin{table*}[h]
\centering
\caption{A--Ci curve data and auxiliary signals for the selected curve used in the Fig.~\ref{Fig:GNN_Edge_Ablation} and Fig.~\ref{Fig:ML_models_test}}
\label{Tab:selected_curve_data}
\begin{tabular}{c|cccc}
\toprule
\textbf{ID} & \textbf{$C_\mathrm{i}$} & \textbf{$A_{\mathrm{net}}$} & \textbf{$s_{A_c}$} & \textbf{$s_{A_j}$} \\
\midrule
0  & 59.60  & 1.14  & 34.23 & 26.99 \\
1  & 158.59 & 9.76  & 62.38 & 74.70 \\
2  & 227.88 & 14.74 & 65.36 & 93.27 \\
3  & 297.17 & 18.07 & 63.58 & 103.11 \\
4  & 336.77 & 20.44 & 65.15 & 112.07 \\
5  & 415.96 & 24.36 & 66.50 & 126.01 \\
6  & 465.45 & 25.02 & 63.30 & 126.03 \\
7  & 554.55 & 27.59 & 62.54 & 134.06 \\
8  & 564.44 & 27.34 & 61.32 & 132.41 \\
9  & 643.64 & 29.48 & 61.34 & 139.52 \\
10 & 712.93 & 29.17 & 57.45 & 135.84 \\
11 & 772.32 & 29.35 & 55.51 & 135.14 \\
12 & 831.72 & 30.53 & 55.71 & 139.19 \\
13 & 930.71 & 29.36 & 50.89 & 132.02 \\
14 & 980.20 & 28.78 & 48.78 & 128.67 \\
\bottomrule
\end{tabular}
\end{table*}

\begin{table*}[!t]
\centering
\caption{Predicted class probabilities for node 5 of the representative A--Ci curve shown in Fig.~\ref{Fig:ML_models_test}. The table reports the probabilities assigned to the three limiting-state classes by the selected graph-based models and compares the predicted label with the true label shown in Fig.~\ref{Fig:GNN_Edge_Ablation}.}
\label{Tab:ProbabilityTable}
\begin{tabular}{cl|cccc|c}
\hline
\multicolumn{2}{c|}{\textbf{Model}} 
& \multicolumn{4}{c|}{\textbf{Predicted}} 
& \multirow{2}{*}{\textbf{True label (Limitation)}} \\
\cline{1-6}
\textbf{Code} 
& \textbf{Name} & \textbf{\(p(A_c)\)} & \textbf{\(p(A_j)\)} & \textbf{\(p(A_p)\)} & \textbf{Label (Limitation)} &  \\ \hline   
\textbf{2-1} & GCN 
& \textbf{0.634} & 0.341 & 0.024 & 0 (Rubisco) & 1 (RuBP) \\
\textbf{3-1} & GCN-U-Net-kNN 
& \textbf{0.545} & 0.422 & 0.031 & 0 (Rubisco) & 1 (RuBP) \\
\textbf{4-1} & GCN-U-Net-ASG 
& 0.237 & \textbf{0.716} & 0.046 & 1 (RuBP) & 1 (RuBP) \\
\textbf{5-1} & GAT-kNN 
& 0.043 & \textbf{0.934} & 0.021 & 1 (RuBP) & 1 (RuBP) \\
\textbf{6-1} & GAT-ASG 
& 0.025 & \textbf{0.921} & 0.053 & 1 (RuBP) & 1 (RuBP) \\
\textbf{7-1} & GAT-U-Net-kNN 
& 0.103 & \textbf{0.882} & 0.014 & 1 (RuBP) & 1 (RuBP) \\
\textbf{8-1} & GAT-U-Net-ASG 
& 0.335 & \textbf{0.647} & 0.017 & 1 (RuBP) & 1 (RuBP) \\
\hline
\end{tabular}
\end{table*}

\begin{table*}[!t]
\centering
\caption{Pairwise Tukey HSD adjusted p-values for F1-score, recall, precision, and accuracy across the selected GNN model configurations.}
{\fontsize{9.5}{9}\selectfont
\begin{tabular}{|l|l|l|l|rrrr|}
\hline
\multicolumn{2}{|c|}{\textbf{Model Group 1}} & \multicolumn{2}{|c|}{\textbf{Model Group 2}} & \multicolumn{4}{|c|}{\textbf{p-value}} \\ \hline
\textbf{Code}  & \textbf{Name} & \textbf{Code}  & \textbf{Name} & \textbf{F1-score} & \textbf{Recall} & \textbf{Precision} & \textbf{Accuracy} \\ \hline
    5-2   & GAT-kNN-$\mathcal{L}_{\mathrm{WCE}}$ & 5-3   & GAT-kNN-$\mathcal{L}_{\mathrm{WF}}$ & 0.800047 & 0.932247 & 0.74386 & 0.628718 \\
    5-2   & GAT-kNN-$\mathcal{L}_{\mathrm{WCE}}$ & 5-1   & GAT-kNN-$\mathcal{L}_{\mathrm{CE}}$ & 0.364223 & 8.1E-08 & 1     & 0.988688 \\
    5-2   & GAT-kNN-$\mathcal{L}_{\mathrm{WCE}}$ & 6-2   & GAT-ASG-$\mathcal{L}_{\mathrm{WCE}}$ & 1.59E-13 & 1.59E-13 & 1.59E-13 & 1.59E-13 \\
    5-2   & GAT-kNN-$\mathcal{L}_{\mathrm{WCE}}$ & 6-3   & GAT-ASG-$\mathcal{L}_{\mathrm{WF}}$ & 1.59E-13 & 1.59E-13 & 1.59E-13 & 1.59E-13 \\
    5-2   & GAT-kNN-$\mathcal{L}_{\mathrm{WCE}}$ & 6-1   & GAT-ASG-$\mathcal{L}_{\mathrm{CE}}$ & 1.59E-13 & 1.59E-13 & 8.58E-09 & 6.03E-07 \\
    5-2   & GAT-kNN-$\mathcal{L}_{\mathrm{WCE}}$ & 7-2   & GAT-U-Net-kNN-$\mathcal{L}_{\mathrm{WCE}}$ & 0.000121 & 0.000151 & 0.000661 & 4.65E-06 \\
    5-2   & GAT-kNN-$\mathcal{L}_{\mathrm{WCE}}$ & 7-3   & GAT-U-Net-kNN-$\mathcal{L}_{\mathrm{WF}}$ & 8.65E-08 & 1.99E-07 & 2.71E-06 & 9.17E-10 \\
    5-2   & GAT-kNN-$\mathcal{L}_{\mathrm{WCE}}$ & 7-1   & GAT-U-Net-kNN-$\mathcal{L}_{\mathrm{CE}}$ & 6.42E-07 & 2.04E-13 & 0.005361 & 0.010949 \\
    5-2   & GAT-kNN-$\mathcal{L}_{\mathrm{WCE}}$ & 8-2   & GAT-U-Net-ASG-$\mathcal{L}_{\mathrm{WCE}}$ & 4.24E-11 & 2.51E-13 & 4.59E-08 & 5.24E-09 \\
    5-2   & GAT-kNN-$\mathcal{L}_{\mathrm{WCE}}$ & 8-3   & GAT-U-Net-ASG-$\mathcal{L}_{\mathrm{WF}}$ & 1.59E-13 & 1.59E-13 & 1.59E-13 & 1.59E-13 \\
    5-2   & GAT-kNN-$\mathcal{L}_{\mathrm{WCE}}$ & 8-1   & GAT-U-Net-ASG-$\mathcal{L}_{\mathrm{CE}}$ & 1.35E-05 & 1.91E-13 & 0.109247 & 0.628718 \\

    5-3   & GAT-kNN-$\mathcal{L}_{\mathrm{WF}}$ & 5-1   & GAT-kNN-$\mathcal{L}_{\mathrm{CE}}$ & 0.999969 & 0.000215 & 0.464141 & 0.049658 \\
    5-3   & GAT-kNN-$\mathcal{L}_{\mathrm{WF}}$ & 6-2   & GAT-ASG-$\mathcal{L}_{\mathrm{WCE}}$ & 1.6E-13 & 1.61E-13 & 1.62E-13 & 1.59E-13 \\
    5-3   & GAT-kNN-$\mathcal{L}_{\mathrm{WF}}$ & 6-3   & GAT-ASG-$\mathcal{L}_{\mathrm{WF}}$ & 1.59E-13 & 1.59E-13 & 1.59E-13 & 1.59E-13 \\
    5-3   & GAT-kNN-$\mathcal{L}_{\mathrm{WF}}$ & 6-1   & GAT-ASG-$\mathcal{L}_{\mathrm{CE}}$ & 2.07E-12 & 1.59E-13 & 0.000229 & 0.009376 \\
    5-3   & GAT-kNN-$\mathcal{L}_{\mathrm{WF}}$ & 7-2   & GAT-U-Net-kNN-$\mathcal{L}_{\mathrm{WCE}}$ & 0.109148 & 0.053407 & 0.318413 & 0.0343 \\
    5-3   & GAT-kNN-$\mathcal{L}_{\mathrm{WF}}$ & 7-3   & GAT-U-Net-kNN-$\mathcal{L}_{\mathrm{WF}}$ & 0.000858 & 0.000436 & 0.013881 & 9.3E-05 \\
    5-3   & GAT-kNN-$\mathcal{L}_{\mathrm{WF}}$ & 7-1   & GAT-U-Net-kNN-$\mathcal{L}_{\mathrm{CE}}$ & 0.00367 & 1.21E-09 & 0.67724 & 0.884834 \\
    5-3   & GAT-kNN-$\mathcal{L}_{\mathrm{WF}}$ & 8-2   & GAT-U-Net-ASG-$\mathcal{L}_{\mathrm{WCE}}$ & 2.26E-06 & 2.27E-09 & 0.000798 & 0.000339 \\
    5-3   & GAT-kNN-$\mathcal{L}_{\mathrm{WF}}$ & 8-3   & GAT-U-Net-ASG-$\mathcal{L}_{\mathrm{WF}}$ & 1.59E-13 & 1.59E-13 & 1.72E-13 & 1.59E-13 \\
    5-3   & GAT-kNN-$\mathcal{L}_{\mathrm{WF}}$ & 8-1   & GAT-U-Net-ASG-$\mathcal{L}_{\mathrm{CE}}$ & 0.028951 & 8.86E-10 & 0.994987 & 1 \\

    5-1   & GAT-kNN-$\mathcal{L}_{\mathrm{CE}}$ & 6-2   & GAT-ASG-$\mathcal{L}_{\mathrm{WCE}}$ & 2.03E-13 & 0.001745 & 1.59E-13 & 1.59E-13 \\
    5-1   & GAT-kNN-$\mathcal{L}_{\mathrm{CE}}$ & 6-3   & GAT-ASG-$\mathcal{L}_{\mathrm{WF}}$ & 1.59E-13 & 4.25E-05 & 1.59E-13 & 1.59E-13 \\
    5-1   & GAT-kNN-$\mathcal{L}_{\mathrm{CE}}$ & 6-1   & GAT-ASG-$\mathcal{L}_{\mathrm{CE}}$ & 1.31E-10 & 2E-11 & 7.5E-10 & 5.02E-10 \\
    5-1   & GAT-kNN-$\mathcal{L}_{\mathrm{CE}}$ & 7-2   & GAT-U-Net-kNN-$\mathcal{L}_{\mathrm{WCE}}$ & 0.424938 & 0.952501 & 0.000112 & 5.35E-09 \\
    5-1   & GAT-kNN-$\mathcal{L}_{\mathrm{CE}}$ & 7-3   & GAT-U-Net-kNN-$\mathcal{L}_{\mathrm{WF}}$ & 0.010408 & 1     & 3.15E-07 & 4.85E-13 \\
    5-1   & GAT-kNN-$\mathcal{L}_{\mathrm{CE}}$ & 7-1   & GAT-U-Net-kNN-$\mathcal{L}_{\mathrm{CE}}$ & 0.034731 & 0.529831 & 0.001102 & 7.24E-05 \\
    5-1   & GAT-kNN-$\mathcal{L}_{\mathrm{CE}}$ & 8-2   & GAT-U-Net-ASG-$\mathcal{L}_{\mathrm{WCE}}$ & 5.63E-05 & 0.603304 & 4.33E-09 & 2.45E-12 \\
    5-1   & GAT-kNN-$\mathcal{L}_{\mathrm{CE}}$ & 8-3   & GAT-U-Net-ASG-$\mathcal{L}_{\mathrm{WF}}$ & 1.59E-13 & 4.9E-08 & 1.59E-13 & 1.59E-13 \\
    5-1   & GAT-kNN-$\mathcal{L}_{\mathrm{CE}}$ & 8-1   & GAT-U-Net-ASG-$\mathcal{L}_{\mathrm{CE}}$ & 0.172433 & 0.493309 & 0.03464 & 0.049658 \\

    6-2   & GAT-ASG-$\mathcal{L}_{\mathrm{WCE}}$ & 6-3   & GAT-ASG-$\mathcal{L}_{\mathrm{WF}}$ & 0.964781 & 0.99957 & 0.891691 & 0.766461 \\
    6-2   & GAT-ASG-$\mathcal{L}_{\mathrm{WCE}}$ & 6-1   & GAT-ASG-$\mathcal{L}_{\mathrm{CE}}$ & 0.99032 & 0.041032 & 0.002182 & 5.33E-07 \\
    6-2   & GAT-ASG-$\mathcal{L}_{\mathrm{WCE}}$ & 7-2   & GAT-U-Net-kNN-$\mathcal{L}_{\mathrm{WCE}}$ & 1.85E-07 & 1.83E-06 & 4.67E-08 & 6.16E-08 \\
    6-2   & GAT-ASG-$\mathcal{L}_{\mathrm{WCE}}$ & 7-3   & GAT-U-Net-kNN-$\mathcal{L}_{\mathrm{WF}}$ & 0.000223 & 0.000905 & 2.23E-05 & 0.000144 \\
    6-2   & GAT-ASG-$\mathcal{L}_{\mathrm{WCE}}$ & 7-1   & GAT-U-Net-kNN-$\mathcal{L}_{\mathrm{CE}}$ & 4.17E-05 & 0.690857 & 2.11E-09 & 1.06E-12 \\
    6-2   & GAT-ASG-$\mathcal{L}_{\mathrm{WCE}}$ & 8-2   & GAT-U-Net-ASG-$\mathcal{L}_{\mathrm{WCE}}$ & 0.028452 & 0.619746 & 0.00067 & 3.77E-05 \\
    6-2   & GAT-ASG-$\mathcal{L}_{\mathrm{WCE}}$ & 8-3   & GAT-U-Net-ASG-$\mathcal{L}_{\mathrm{WF}}$ & 0.999327 & 0.623279 & 1     & 1 \\
    6-2   & GAT-ASG-$\mathcal{L}_{\mathrm{WCE}}$ & 8-1   & GAT-U-Net-ASG-$\mathcal{L}_{\mathrm{CE}}$ & 2.2E-06 & 0.724754 & 5.24E-12 & 1.59E-13 \\

    6-3   & GAT-ASG-$\mathcal{L}_{\mathrm{WF}}$ & 6-1   & GAT-ASG-$\mathcal{L}_{\mathrm{CE}}$ & 0.301884 & 0.328877 & 8.71E-07 & 5.05E-12 \\
    6-3   & GAT-ASG-$\mathcal{L}_{\mathrm{WF}}$ & 7-2   & GAT-U-Net-kNN-$\mathcal{L}_{\mathrm{WCE}}$ & 3.77E-11 & 1.69E-08 & 1.54E-12 & 5.15E-13 \\
    6-3   & GAT-ASG-$\mathcal{L}_{\mathrm{WF}}$ & 7-3   & GAT-U-Net-kNN-$\mathcal{L}_{\mathrm{WF}}$ & 1.84E-07 & 1.96E-05 & 2.36E-09 & 5.78E-09 \\
    6-3   & GAT-ASG-$\mathcal{L}_{\mathrm{WF}}$ & 7-1   & GAT-U-Net-kNN-$\mathcal{L}_{\mathrm{CE}}$ & 2.34E-08 & 0.169936 & 1.96E-13 & 1.59E-13 \\
    6-3   & GAT-ASG-$\mathcal{L}_{\mathrm{WF}}$ & 8-2   & GAT-U-Net-ASG-$\mathcal{L}_{\mathrm{WCE}}$ & 0.000111 & 0.132792 & 1.81E-07 & 1.01E-09 \\
    6-3   & GAT-ASG-$\mathcal{L}_{\mathrm{WF}}$ & 8-3   & GAT-U-Net-ASG-$\mathcal{L}_{\mathrm{WF}}$ & 0.999996 & 0.982513 & 0.77473 & 0.605726 \\
    6-3   & GAT-ASG-$\mathcal{L}_{\mathrm{WF}}$ & 8-1   & GAT-U-Net-ASG-$\mathcal{L}_{\mathrm{CE}}$ & 6.87E-10 & 0.191109 & 1.59E-13 & 1.59E-13 \\

    6-1   & GAT-ASG-$\mathcal{L}_{\mathrm{CE}}$ & 7-2   & GAT-U-Net-kNN-$\mathcal{L}_{\mathrm{WCE}}$ & 8.35E-05 & 1.6E-13 & 0.577467 & 1 \\
    6-1   & GAT-ASG-$\mathcal{L}_{\mathrm{CE}}$ & 7-3   & GAT-U-Net-kNN-$\mathcal{L}_{\mathrm{WF}}$ & 0.02328 & 7.07E-12 & 0.997204 & 0.994357 \\
    6-1   & GAT-ASG-$\mathcal{L}_{\mathrm{CE}}$ & 7-1   & GAT-U-Net-kNN-$\mathcal{L}_{\mathrm{CE}}$ & 0.006624 & 9.86E-06 & 0.240418 & 0.597635 \\
    6-1   & GAT-ASG-$\mathcal{L}_{\mathrm{CE}}$ & 8-2   & GAT-U-Net-ASG-$\mathcal{L}_{\mathrm{WCE}}$ & 0.483289 & 5.87E-06 & 1     & 0.99962 \\
    6-1   & GAT-ASG-$\mathcal{L}_{\mathrm{CE}}$ & 8-3   & GAT-U-Net-ASG-$\mathcal{L}_{\mathrm{WF}}$ & 0.65854 & 0.983792 & 0.005359 & 1.92E-06 \\
    6-1   & GAT-ASG-$\mathcal{L}_{\mathrm{CE}}$ & 8-1   & GAT-U-Net-ASG-$\mathcal{L}_{\mathrm{CE}}$ & 0.000645 & 1.27E-05 & 0.017691 & 0.009376 \\

    7-2   & GAT-U-Net-kNN-$\mathcal{L}_{\mathrm{WCE}}$ & 7-3   & GAT-U-Net-kNN-$\mathcal{L}_{\mathrm{WF}}$ & 0.964892 & 0.979788 & 0.992393 & 0.943355 \\
    7-2   & GAT-U-Net-kNN-$\mathcal{L}_{\mathrm{WCE}}$ & 7-1   & GAT-U-Net-kNN-$\mathcal{L}_{\mathrm{CE}}$ & 0.99686 & 0.014185 & 0.999996 & 0.84194 \\
    7-2   & GAT-U-Net-kNN-$\mathcal{L}_{\mathrm{WCE}}$ & 8-2   & GAT-U-Net-ASG-$\mathcal{L}_{\mathrm{WCE}}$ & 0.267522 & 0.020056 & 0.771311 & 0.98844 \\
    7-2   & GAT-U-Net-kNN-$\mathcal{L}_{\mathrm{WCE}}$ & 8-3   & GAT-U-Net-ASG-$\mathcal{L}_{\mathrm{WF}}$ & 1.03E-09 & 5.58E-12 & 1.75E-07 & 2.38E-07 \\
    7-2   & GAT-U-Net-kNN-$\mathcal{L}_{\mathrm{WCE}}$ & 8-1   & GAT-U-Net-ASG-$\mathcal{L}_{\mathrm{CE}}$ & 0.999999 & 0.011878 & 0.952172 & 0.0343 \\

    7-3   & GAT-U-Net-kNN-$\mathcal{L}_{\mathrm{WF}}$ & 7-1   & GAT-U-Net-kNN-$\mathcal{L}_{\mathrm{CE}}$ & 1     & 0.417639 & 0.884153 & 0.058093 \\
    7-3   & GAT-U-Net-kNN-$\mathcal{L}_{\mathrm{WF}}$ & 8-2   & GAT-U-Net-ASG-$\mathcal{L}_{\mathrm{WCE}}$ & 0.985392 & 0.488711 & 0.999871 & 1 \\
    7-3   & GAT-U-Net-kNN-$\mathcal{L}_{\mathrm{WF}}$ & 8-3   & GAT-U-Net-ASG-$\mathcal{L}_{\mathrm{WF}}$ & 3.1E-06 & 1.94E-08 & 6.81E-05 & 0.00042 \\
    7-3   & GAT-U-Net-kNN-$\mathcal{L}_{\mathrm{WF}}$ & 8-1   & GAT-U-Net-ASG-$\mathcal{L}_{\mathrm{CE}}$ & 0.998688 & 0.383606 & 0.286284 & 9.3E-05 \\

    7-1   & GAT-U-Net-kNN-$\mathcal{L}_{\mathrm{CE}}$ & 8-2   & GAT-U-Net-ASG-$\mathcal{L}_{\mathrm{WCE}}$ & 0.909417 & 1     & 0.411056 & 0.126284 \\
    7-1   & GAT-U-Net-kNN-$\mathcal{L}_{\mathrm{CE}}$ & 8-3   & GAT-U-Net-ASG-$\mathcal{L}_{\mathrm{WF}}$ & 4.51E-07 & 0.002928 & 8.57E-09 & 4.69E-12 \\
    7-1   & GAT-U-Net-kNN-$\mathcal{L}_{\mathrm{CE}}$ & 8-1   & GAT-U-Net-ASG-$\mathcal{L}_{\mathrm{CE}}$ & 0.999988 & 1     & 0.998729 & 0.884834 \\

    8-2   & GAT-U-Net-ASG-$\mathcal{L}_{\mathrm{WCE}}$ & 8-3   & GAT-U-Net-ASG-$\mathcal{L}_{\mathrm{WF}}$ & 0.001103 & 0.001954 & 0.001757 & 0.000116 \\
    8-2   & GAT-U-Net-ASG-$\mathcal{L}_{\mathrm{WCE}}$ & 8-1   & GAT-U-Net-ASG-$\mathcal{L}_{\mathrm{CE}}$ & 0.567916 & 1     & 0.044278 & 0.000339 \\
    8-3   & GAT-U-Net-ASG-$\mathcal{L}_{\mathrm{WF}}$ & 8-1   & GAT-U-Net-ASG-$\mathcal{L}_{\mathrm{CE}}$ & 1.63E-08 & 0.00357 & 2.38E-11 & 1.59E-13 \\
    \hline
    \end{tabular}}
  \label{Tab:p-Value}%
\end{table*}%

\end{document}